\pdfoutput=1

\documentclass[11pt,table]{article}

\usepackage[final]{acl}

\usepackage{times}
\usepackage{latexsym}

\usepackage[T1]{fontenc}

\usepackage[utf8]{inputenc}

\usepackage{microtype}

\usepackage{inconsolata}

\usepackage{graphicx}
\usepackage{booktabs}
\usepackage{multirow}
\usepackage{hyperref}
\usepackage{placeins}

\usepackage{float}  
\usepackage{lipsum}
\usepackage{dblfloatfix}
\usepackage{enumitem}
\DeclareRobustCommand{\name}{\textsc{ProReason}}
\title{\name{}: Multi-Modal Proactive Reasoning\\
with Decoupled Eyesight and Wisdom}

\author{
\bf Jingqi Zhou$^{\heartsuit}$, Sheng Wang$^{\heartsuit}$\textsuperscript{\dag}, Jingwei Dong$^{\heartsuit}$, Kai Liu$^{\heartsuit}$, \\
\bf Lei Li$^{\heartsuit}$, Jiahui Gao$^{\heartsuit}$, Jiyue Jiang$^{\spadesuit}$, Lingpeng Kong$^{\heartsuit}$, Chuan Wu$^{\heartsuit}$\textsuperscript{\dag} \\
$^{\heartsuit}$ The University of Hong Kong, $^{\spadesuit}$ The Chinese University of Hong Kong,\\
{\tt \{u3011211, u3009618, u3013005, u3638070\}@connect.hku.hk,}\\
{\tt nlp.lilei@gmail.com, sumiler@connect.hku.hk,}\\
{\tt jiangjy@link.cuhk.edu.hk, \{lpk, cwu\}@cs.hku.hk}
}

\begin{document}
\maketitle
\begin{abstract}
Large vision-language models (LVLMs) have witnessed significant progress on visual understanding tasks. 
However, they often prioritize language knowledge over image information on visual reasoning tasks, incurring performance degradation.
To tackle this issue,  we first identify the drawbacks of existing solutions (\textit{i.e.}, limited multi-modal reasoning capacities, and insufficient and irrelevant visual descriptions).
We then decompose visual reasoning process into two stages: proactive visual perception (\textit{i.e.}, eyesight) and textual reasoning (\textit{i.e.}, wisdom), and introduce a novel visual reasoning framework named \name{}. 
This framework features decoupled vision-reasoning capabilities and multi-run proactive perception.
Briefly, given a multi-modal question, \name{} iterates 
proactive information collection and reasoning
until the answer can be concluded with necessary and sufficient visual descriptions.
Notably, the disassociation of capabilities allows seamless integration of existing large language models (LLMs) to compensate for the reasoning deficits of LVLMs.
Our extensive experiments demonstrate that \name{} outperforms existing multi-step reasoning frameworks on various benchmarks for both open-source and closed-source models, with the average performance gain reaching 13.2\%.
Besides, the integration of LLMs allows \name{} to produce high-quality visual reasoning data, which empowers \name{}-distilled models (\textit{i.e.}, \texttt{\name{}-VL} and \texttt{\name{}-Q3}) to achieve superior performance in downstream tasks.
Our insights into existing solutions and the decoupled perspective for feasible integration of LLMs illuminate future research on visual reasoning techniques, especially LLM-assisted ones. The code is available at \url{https://github.com/lian-tian-mo-zun/Pro_Reason}.
\end{abstract}

\section{Introduction}
\vspace{-0.5em}
In recent years, large language models (LLMs)~\citep{qwen2.5, llama3, GeminiAFamily_Team2023,  Mistral7B_Jiang2023} have experienced explosive growth in their capabilities, driving significant advancements across various fields~\citep{CharacterLlmA_Shao2023,ds_code,ds_math}. 
This progress has also sparked interest in developing large vision-language models (LVLMs)~\citep{xiaomi2025mimo, Qwen2.5-VL, HowFarAre_Chen2024, QwenVlA_Bai2023}, which, like LLaVA~\citep{li2024llava}, have achieved remarkable performance in multi-modal understanding tasks. 
However, state-of-the-art (SOTA) LVLMs still struggle to handle visual understanding with textual reasoning simultaneously due to inherent modality differences. 
For example, \citet{ghosh2024vdgd} demonstrate that LVLMs often rely more on their prior language knowledge, neglecting visual information in multi-modal reasoning tasks, such as visual chart understanding and math reasoning, resulting in performance degradation. Figure~\ref{fig:case}.b illustrates a typical case of this issue, where the reasoning process remains irrelevant to the image.

To address the challenges, several visual reasoning frameworks have been proposed. Specifically,  \cite{ghosh2024vdgd} and \cite{mitra2024compositional} convert visual information in images into textual descriptions to aid LVLMs in reasoning. 
However, their visual extraction process is not targeted at a given question (\textit{i.e.}, question-agnostic), termed as "passive", and omits reasoning mechanisms to infer extra information for better descriptions(\textit{i.e.}, reasoning-free).
These limitations result in irrelevant or inadequate information, ultimately degrading performance.
Furthermore, these frameworks are powered by a single LVLM, leading to a reasoning process that conflates visual understanding with textual reasoning abilities, failing to mitigate the challenge faced by LVLMs in effectively managing both capabilities.

\begin{figure*}[!t]
    \caption {Overview and comparison of \name{}, VDGD and ReAct. Unlike existing works (\textit{e.g.}, VDGD and ReAct), our proposed method decouples visual perception and textual reasoning while allowing the model to actively acquire necessary information from the images, achieving superior performance.}
    \label{fig:method}
    \begin{minipage}{\textwidth}
        \centering
        \includegraphics[width=1.0\textwidth]{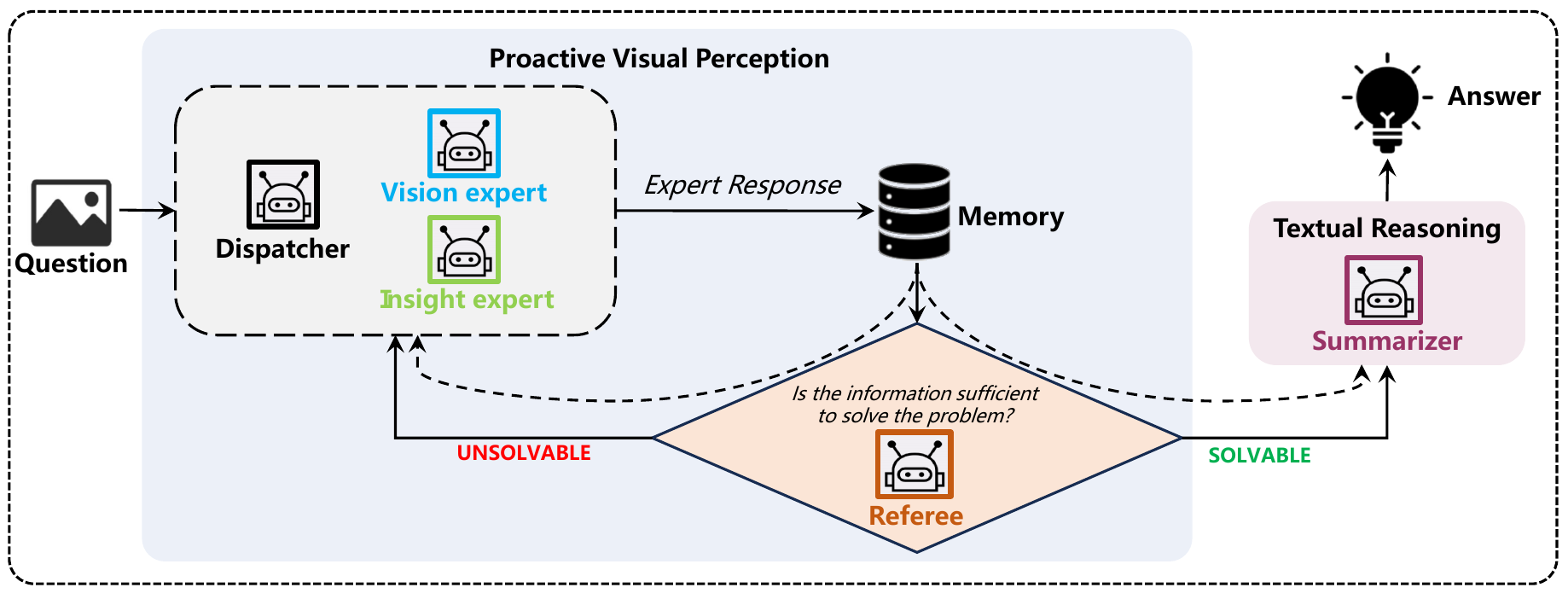}
        \caption*{(1) \name{}}
    \end{minipage}

    \begin{minipage}[t]{0.51\textwidth} 
        \centering
        \includegraphics[width=1.0\textwidth]{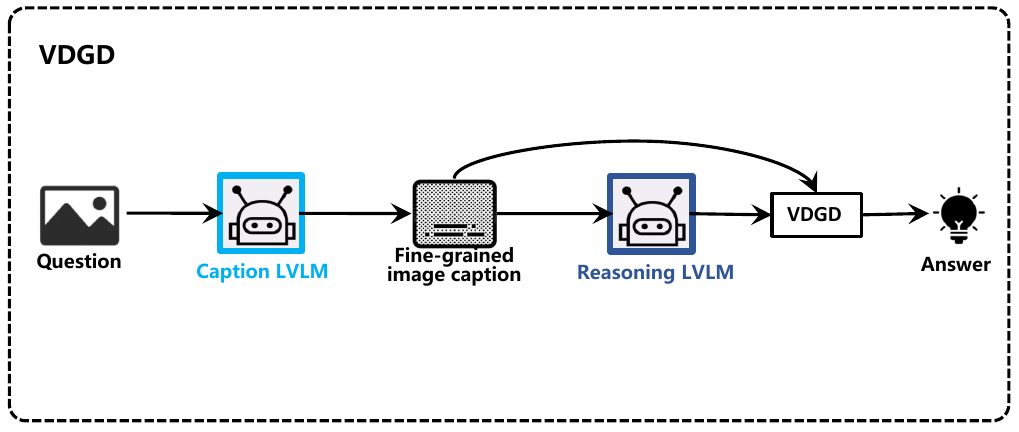}
        \caption*{(2) Passive Visual Reasoning Enhancement Technique}
    \end{minipage}
    \hfill
    \begin{minipage}[t]{0.48\textwidth} 
        \centering
        \includegraphics[width=1.0\textwidth]{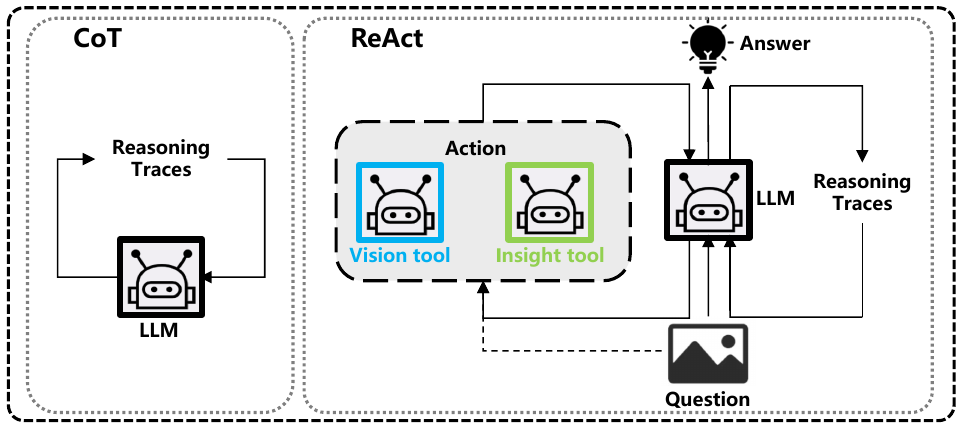}
        \caption*{(3) Multi-step Reasoning Framework for LLMs}
    \end{minipage}
\end{figure*}

To resolve these problems, we propose \name{}, a multi-modal reasoning framework featuring decoupled vision-reasoning capabilities. 
As illustrated in Figure~\ref{fig:method}, we decouple multi-modal reasoning capacity into two sub-tasks: proactive visual perception (\textit{i.e.}, eyesight) and textual reasoning (\textit{i.e.}, wisdom). The former extracts visual information in a question-oriented and reasoning-involved manner, while the latter integrates all information to draw final conclusions.
Specifically, during the visual perception stage, a Dispatcher first selectively engages a Vision Expert to capture additional visual information, or an Insight Expert to derive intermediate inferences.
A Referee then determines whether sufficient information is gathered to proceed to the reasoning stage, where a Summarizer produces the final answer.
Unlike passive methods, all sub-agents operate based on the given question and known information, effectively avoiding irrelevant information redundancy or insufficiency.
Notably, decoupled vision-reasoning eliminates the need for LVLMs to handle vision-irrelevant roles (\textit{i.e.}, Dispatcher, Insight Expert, Referee, and Summarizer), enabling seamless integration of existing LLMs with proven strong reasoning abilities~\citep{chang2024survey}, thereby alleviate the limitations of LVLMs. 
In addition, the high-quality reasoning data generated by LLM-assisted 
\name{} can be effectively distilled 
into downstream models for inherent performance improvement.

Empirically, we evaluate \name{} across multiple challenging visual reasoning benchmarks with both open-source and closed-source models.
Extensive experiments demonstrate that \name{} exhibits significant advantages in two key aspects: (1) As a visual reasoning framework, \name{} achieves consistent and substantial performance improvements across multiple benchmarks, with the average performance gain reaching 13.2\%, validating the effectiveness of its decoupled vision-reasoning architecture and proactive visual feature extraction mechanism; 
(2) \name{} effectively integrates existing LLMs to generate high-quality visual reasoning process, empowering \name{}-distilled models (\textit{i.e.}, \texttt{\name{}-VL} and \texttt{\name{}-Q3}) with superior visual reasoning 
capabilities.
The above results, coupled with the ablation study in Section~\ref{sec: more experiment}, 
demonstrate the substantial advantages of decoupled vision-reasoning, while highlighting the potential of LLM-assisted LVLM reasoning and distillation strategies.

The main contributions of this work are threefold:
\begin{itemize}[itemsep=2pt, topsep=0pt]
    \item We propose a novel multi-modal reasoning framework named \name{}, featuring decoupled vision-reasoning and iterative proactive perception capabilities, effectively mitigating the drawbacks of previous methods. 
    \item Extensive experiments consistently highlight the significant superiority of \name{} and necessity of each component across multiple visual reasoning tasks and model series, illuminating the great potential of LLM-assisted LVLM reasoning.
    \item \name{}-distilled models also exhibits remarkable enhancements over vanilla counterparts, showcasing the feasibility of LLM-assisted LVLM improvement in the future.
\end{itemize}

\section{Preliminary Observations} \label{sec: passive analysis}
\citet{ghosh2024vdgd} demonstrate that the limited multimodal reasoning abilities of LVLMs lead to an overreliance on linguistic priors, thus neglecting visual inputs and ultimately degrading their performance. Their Visual Description Grounded Decoding (VDGD) mitigates visual oversight by converting images into comprehensive textual descriptions to inform reasoning processes.
However, \textbf{such passive visual reasoning techniques suffer insufficient and irrelevant visual information}.
To support this claim, we generate fine-grained image captions using \texttt{GPT-4o-mini}\footnote{\scriptsize \url{https://openai.com/index/gpt-4o-mini-advancing-cost-efficient-intelligence}} with the prompt shown in Figure~\ref{fig:prompt_baseline}. 
We then incorporate these captions into the prompts for LVLMs to facilitate the reasoning process. 
We analyze the performance of this approach on the challenging multi-modal MMMU dataset, which requires college-level knowledge and fine-grained reasoning, using recent open-source LVLMs listed in Section~\ref{sec: base model}
As shown in Table~\ref{table:motivation}, 
while these image descriptions improve the performance of LVLMs, the gains are marginal, consistently amounting to less than 1\%. This underscores the limited utility of captions generated by passive methods.

For further demonstration, inspired by \citet{liu2023llava}, we use the prompt instructions in Figure~\ref{fig:score} to instruct \texttt{GPT-4} to analyze the generated captions along three dimensions: Detail Level, Question Relevance, and Reasoning Effective Info Inclusion, measuring the richness of detail, relevance to the given question, and the inclusion of information that is essential for reasoning, respectively.
Meanwhile, since the reasoning process of \texttt{GPT-4o-mini} on MMMU contains key information necessary for solving the problems, we use it as a reference answer to aid evaluation.
As shown in Table~\ref{table:cap_evalue}, the captions for correct responses of \texttt{Llama3-LLaVA-NeXT-8B} receive higher scores across all three criteria, highlighting the importance of better captions for multi-modal reasoning. Additionally, all captions score significantly lower in the Question Relevance and Reasoning Effective Info Inclusion dimensions than the Detail Level dimension, indicating that \textbf{while the captions are detailed, they often lack relevance to the questions.}
Figure~\ref{fig:case} shows a case where \texttt{Llama3-LLaVA-NeXT-8B} utilizes fine-grained captions to solve a question from the MMMU benchmark. 
As illustrated, although the caption exhaustively describes the image content, it incorrectly describes the wires in the image as octagons, and misses information about the locations of these wires. This information is irrelevant to the target question,  thus offering minimal assistance to LVLMs. 
In summary, our analysis highlights the drawbacks of passive visual reasoning enhancement techniques in terms of information insufficiency and redundancy, due to their question-agnostic property.

\section{Method} \label{sec: method} 
As depicted in Figure~\ref{fig:method} and exemplified in Figure~\ref{fig:full case}, \name{} presents an innovative decoupling of the visual reasoning process into two distinct phases framed through the lens of LVLM capabilities: Proactive Visual Perception (\textit{i.e.}, eyesight) and Textual Reasoning (\textit{i.e.}, wisdom). The entire workflow consists of five functionally distinct yet inter-cooperative sub-agents, along with a Memory component, facilitating multi-modal reasoning performance.

\subsection{Proactive Visual Perception}
Proactive Visual Perception is the core of question-oriented visual information extraction, driven by four sub-agents: Dispatcher, Vision Expert, Insight Expert and Referee. The Dispatcher breaks down the original question, selectively directing the Vision Expert to capture specific visual information, or instructing the Insight Expert to analyze known information to derive more. 
The responses from both experts are stored in a textual Memory component. The Referee then evaluates whether the information stored in Memory is sufficient to answer the original question.

Formally, given an image $I$ and its corresponding textual question $Q$, the Dispatcher decides to consult the Vision Expert or Insight Expert, based on the analysis of $Q$ and the known information in the Memory (if not empty). The Dispatcher then generates a query $q$ for the chosen expert. If the Vision Expert is selected, it takes the image $I$ and query $q$ as input, and generates an answer $A_{vision}$, which is then stored in the Memory. 
When the Insight Expert is selected, it provides a response consisting of the inference process and the final answer $A_{insight}$ based on the query $q$ and known information in the Memory, before only $A_{insight}$ is stored in Memory. 
The Referee then evaluates the available information in the Memory concerning the question $Q$. If the Memory contains adequate information to answer the question $Q$, the Referee outputs the identifier ``SOLVABLE''; otherwise, it outputs ``UNSOLVABLE''. If the Referee's output is SOLVABLE, the workflow precedes to the Textual Reasoning phase. Conversely, if the output is UNSOLVABLE, the above process will be re-executed to gather more necessary information. 

In the Proactive Visual Perception phase, the Dispatcher collaborates with the Vision Expert and Insight Expert to achieve question-oriented and reasoning-involved visual information extraction, while the Referee ensures informational completeness and prevents omissions. These four sub-agents work closely together, thereby overcoming the limitations of passive methods.
Notably, the Memory component allows \name{} to keep compact information, and avoids lengthy reasoning traces like CoT~\citep{wei2022chain} and ReAct~\citep{yao2022react}, thereby suffering less from redundant information (for a detailed analysis, refer to Section~\ref{sec:cost} and Table~\ref{table: cost}).

\subsection{Textual Reasoning Step}
The Textual Reasoning step focuses on integrating the available information in the Memory, and providing the final answer to the question $Q$.
This step is mainly powered by a sub-agent called Summarizer.
Once the Referee confirms that the Memory contains sufficient information to address question $Q$ and outputs the SOLVABLE identifier, the Summarizer will be invoked to perform a detailed reasoning and generate a final answer based on the question $Q$ and Memory. This final answer is then evaluated using performance metrics, while the Summarizer's high-quality reasoning outputs can be utilized to train downstream models, enhancing their visual reasoning capabilities.

\subsection{Advantages of \name{}} \label{sec:advantages}

\paragraph{Decoupled Visual Reasoning.}
In \name{}, the multi-modal reasoning process is decomposed into visual perception and textual reasoning stages, each executed by separate agents. These agents are then effectively organized through a designated pipeline, significantly enhancing the ability of LVLMs to tackle visual reasoning tasks.

\paragraph{LLM-assisted multi-modal reasoning.}
Decoupled vision-reasoning eliminates the necessity for LVLMs to manage vision-irrelevant sub-agents, allowing the seamless integration of existing LLMs with established strong reasoning abilities, thereby endowing \name{} with superior visual reasoning performance.

\paragraph{Downstream task model enhancement.}
By generating high-quality data, \name{} effectively distills its superior capabilities into downstream task models, enabling them to demonstrate exceptional performance in complex visual reasoning tasks.

\paragraph{Reduced information Mission or Redundancy.}
Through sub-agent collaboration in Proactive Visual Perception, \name{} extracts essential visual details via question-oriented manner, preventing information omission or redundancy. Meanwhile, the Memory component retains only Vision Expert's observations and Insight Expert's conclusions, creating compact representation that minimizes irrelevant information. These advantages effectively reduce the overhead when \name{} functions as a reasoning framework.

\section{Experiments} \label{sec: exp}

In this section, we first evaluate the performance of our \name{} framework against recent baselines on multiple benchmarks, followed by an in-depth ablation analysis of different components.

\subsection{General Setup} \label{sec: setup}
\paragraph{Datasets.}
To comprehensively validate the performance of our framework, we conduct experiments across four benchmarks: Multi-modal Large Language Model Evaluation (MME)~\citep{yin2023survey} \footnote{Due to our emphasis on visual reasoning, we select the cognition-relevant tasks, including Commonsense Reasoning, Numerical Calculation, Text Translation, and Code Reasoning. To facilitate the comparison across different benchmarks, the results for MME benchmark are calculated by the percentage of correct answers out of the total answers. 
}
, Massive Multi-discipline Multi-modal Understanding and Reasoning (MMMU)~\citep{yue2023mmmu}, MathVista~\citep{wang2024measuring}, and HallusionBench~\citep{liu2023hallusionbench}. All of them require visual reasoning capabilities to complete the tasks correctly, and are introduced briefly in Section~\ref{sec: Dataset}.

\paragraph{Base Models.} \label{sec: base model}
We employ \texttt{GPT-4o-mini}, \texttt{Llama3-LLaVA-NeXT-8B} and \texttt{Qwen2.5-VL-7B- Instruct}~\citep{Qwen2.5-VL}\footnote{The model is deployed using CUDA \url{https://developer.nvidia.com/cuda-toolkit} on a NVIDIA A100 graphics cards.} to drive \name{}, and \texttt{Qwen2.5-72B-Instruct}~\citep{qwen2.5} and \texttt{Qwen3-32B}~\citep{qwen3} are selected to drive the text-only sub-agents, enabling LLM-assisted visual reasoning.
In addition \texttt{Qwen2.5-VL-7B-Instruct} is also utilized as a downstream task model, which is trained with the data generated by \name{}.
These models are chosen for their leading performance and popularity.

\paragraph{Baselines.} \label{sec: Baselines}
In addition to the basic method where models are instructed to answer questions directly, we compare \name{} with the following peer methods. 
First, to explore the effectiveness of directly migrating LLM solutions to LVLMs, we select two multi-step reasoning frameworks from LLMs: Chain of Thought (CoT)~\citep{wei2022chain} and ReAct~\citep{yao2022react}. 
Second, to evaluate the advantages of proactive information extraction in \name{}, we examine two passive visual reasoning frameworks, VDGD~\citep{ghosh2024vdgd} and CCoT~\citep{mitra2024compositional}, which assist LVLM reasoning by extracting image information into text. 
Additionally, we consider \texttt{R1-Onevision-7B}~\citep{yang2025r1}, an LVLM based on \texttt{Qwen2.5-VL-7B-Instruct} with deep thinking capabilities similar to \texttt{Deekseek-R1}~\citep{guo2025deepseek} and \texttt{OpenAI o1}\footnote{\url{https://openai.com/index/introducing-openai-o1-preview/}}. We compare it with \texttt{\name{}-Q3}, which also exhibits deep thinking abilities.
Further details on the baselines and implementation are provided in Section~\ref{sec: baseline_app}.

\paragraph{Implementation Details.} \label{sec: implementation Details}
The prompt templates for all methods are shown in Figures~\ref{fig:prompt_baseline}, \ref{fig:prompt_dis} and \ref{fig:prompt_else}.
Specifically for \name{}, to prevent infinite loops, if the Dispatcher selects the Vision Expert or Insight Expert to obtain information up to 5 consecutive times, and the Referee still determines that the existing knowledge in Memory remains insufficient to resolve the question, the Memory will be cleared to restart the information acquisition process. If the system fail to break the loop after 5 attempts, the proactive visual perception phase will be immediately terminated. Subsequently, the Summarizer will generate the final answer based on the available information in Memory from the last attempted iteration.
This setup, refined through multiple trials, is the most effective. 

\begin{table*}[!t]
    \caption{
    Performance of multiple approaches with three base models across four visual reasoning benchmarks.
    ``Hallu.'' is the abbreviation of HallusionBench.
    The abbreviations \texttt{4o-mini}, \texttt{Qwen72B}, and \texttt{Qwen3} refer to \texttt{GPT-4o-mini}, \texttt{Qwen2.5-72B-Instruct}, and \texttt{Qwen3-32B}, respectively. 
    "Assisted" stands for LLM-assisted reasoning, which involves replacing the textual sub-agents within frameworks with corresponding LLMs, as detailed in Table~\ref{table:text agent set}. 
    Based on the performance of the direct method, red and blue signify the improvement and degradation, respectively.}
    \label{table:main}
    \resizebox{\textwidth}{!}{
    \centering
    \renewcommand{\arraystretch}{0.8} 
    \begin{tabular}{c l c c c c c}
        \toprule
             \multirow{2}{*}{Model} & \multirow{2}{*}{Method} & \multicolumn{5}{c}{Dataset} \\
             \cmidrule{3-7}& & MME & MMMU & MathVista & Hallu. & Average \\
        \midrule
        \multirow{6}{*}{\begin{tabular}{c} \texttt{GPT-4o-mini} \end{tabular}}
             & Direct & 79.2 & 48.4 & 53.0 & 56.0 & 59.2 \\
             & VDGD & 82.3 \textcolor{red}{(+3.1)} & 51.4 \textcolor{red}{(+3.0)} & 51.2 \textcolor{blue}{(-1.8)} & 52.4 \textcolor{blue}{(-3.6)} & 59.3 \textcolor{red}{(+0.1)} \\
             & CCoT & 80.8 \textcolor{red}{(+1.6)} & 54.2 \textcolor{red}{(+5.8)} & 53.6 \textcolor{red}{(+0.6)} & 56.7 \textcolor{red}{(+0.7)} & 61.3 \textcolor{red}{(+2.1)} \\
             & CoT & 87.8 \textcolor{red}{(+8.6)} & 58.5 \textcolor{red}{(+10.1)} & 53.8 \textcolor{red}{(+0.8)} & 56.3 \textcolor{red}{(+0.3)} & 64.1 \textcolor{red}{(+4.9)} \\
             & ReAct & 87.3 \textcolor{red}{(+8.1)} & 54.8 \textcolor{red}{(+6.4)} & 49.3 \textcolor{blue}{(-3.7)} & 51.1 \textcolor{blue}{(-4.9)} & 60.6 \textcolor{red}{(+1.4)} \\
             &\cellcolor[HTML]{ECF4FF} \name{} &\cellcolor[HTML]{ECF4FF} 91.9 \textcolor{red}{(+12.7)} &\cellcolor[HTML]{ECF4FF} 61.6 \textcolor{red}{(+13.2)} &\cellcolor[HTML]{ECF4FF} 54.9 \textcolor{red}{(+1.9)} &\cellcolor[HTML]{ECF4FF} 59.9 \textcolor{red}{(+3.9)} &\cellcolor[HTML]{ECF4FF} 67.1 \textcolor{red}{(+7.9)} \\
        \midrule
            \multirow{11}{*}{\texttt{Llama3-LLaVA-NeXT-8B}} 
             & Direct & 61.5 & 41.8 & 37.1 & 45.8 & 46.6 \\
             & VDGD & 68.8 \textcolor{red}{(+7.3)} & 42.3 \textcolor{red}{(+0.5)} & 36.1 \textcolor{blue}{(-1.0)} & 44.2 \textcolor{blue}{(-1.6)} & 47.8 \textcolor{red}{(+1.2)}\\
             & CCoT & 68.9 \textcolor{red}{(+7.4)} & 40.5 \textcolor{blue}{(-1.3)} & 36.8 \textcolor{blue}{(-0.3)} & 37.4 \textcolor{blue}{(-8.4)} & 45.9 \textcolor{blue}{(-0.7)} \\
             & CoT & 58.8 \textcolor{blue}{(-2.7)} & 41.5 \textcolor{blue}{(-0.3)}& 35.9 \textcolor{blue}{(-1.2)} & 43.1 \textcolor{blue}{(-2.7)} & 44.8 \textcolor{blue}{(-1.8)}\\
             \cmidrule{2-7}
             &  ReAct &  68.5 \textcolor{red}{(+7.0)} &  46.7 \textcolor{red}{(+4.9)} &  31.7 \textcolor{blue}{(-5.4)} &  43.6 \textcolor{blue}{(-2.2)} &  47.6 \textcolor{red}{(+1.0)} \\
             & + \texttt{4o-mini} Assisted & 73.6 \textcolor{red}{(+12.1)} & 48.4 \textcolor{red}{(+6.6)} & 36.2 \textcolor{blue}{(-0.9)} & 46.7 \textcolor{red}{(+0.9)} & 51.2 \textcolor{red}{(+4.6)} \\ 
             & + \texttt{Qwen72B} Assisted & 71.0 \textcolor{red}{(+9.5)} & 50.4 \textcolor{red}{(+8.6)} & 34.6 \textcolor{blue}{(-2.5)} & 40.4 \textcolor{blue}{(-5.4)} & 49.1 \textcolor{red}{(+2.5)} \\ 
             \cmidrule{2-7}
             &\cellcolor[HTML]{ECF4FF} \name{} &\cellcolor[HTML]{ECF4FF} 71.5 \textcolor{red}{(+10.0)} &\cellcolor[HTML]{ECF4FF} 50.5 \textcolor{red}{(+8.7)} &\cellcolor[HTML]{ECF4FF} 38.8 \textcolor{red}{(+1.7)} &\cellcolor[HTML]{ECF4FF} 50.9 \textcolor{red}{(+5.1)} &\cellcolor[HTML]{ECF4FF} 52.9 \textcolor{red}{(+6.3)}\\ 
             &\cellcolor[HTML]{ECF4FF}+ \texttt{4o-mini} Assisted &\cellcolor[HTML]{ECF4FF} 84.7 \textcolor{red}{(+23.2)} &\cellcolor[HTML]{ECF4FF} 54.5 \textcolor{red}{(+12.7)} &\cellcolor[HTML]{ECF4FF} 41.7 \textcolor{red}{(+4.6)} &\cellcolor[HTML]{ECF4FF} 53.1 \textcolor{red}{(+7.3)} &\cellcolor[HTML]{ECF4FF} 58.5 \textcolor{red}{(+11.9)} \\ 
             &\cellcolor[HTML]{ECF4FF}+ \texttt{Qwen72B} Assisted &\cellcolor[HTML]{ECF4FF} 81.3 \textcolor{red}{(+19.8)} &\cellcolor[HTML]{ECF4FF} 56.8 \textcolor{red}{(+15.0)} &\cellcolor[HTML]{ECF4FF} 48.8 \textcolor{red}{(+11.7)} &\cellcolor[HTML]{ECF4FF} 52.3 \textcolor{red}{(+6.5)} &\cellcolor[HTML]{ECF4FF} 59.8 \textcolor{red}{(+13.2)} \\ 
        \midrule
            \multirow{10}{*}{\texttt{Qwen2.5-VL-7B-Instruct}} 
             & Direct & 74.2 & 51.8 & 63.3 & 53.8 & 60.8 \\
             & VDGD & 74.6 \textcolor{red}{(+0.4)} & 52.3 \textcolor{red}{(+0.5)} & 62.1 \textcolor{blue}{(-1.2)} & 53.9 \textcolor{red}{(+0.1)} & 60.7 \textcolor{blue}{(-0.1)} \\
             & CCoT & 82.7 \textcolor{red}{(+8.5)} & 52.2 \textcolor{red}{(+0.4)} & 61.7 \textcolor{blue}{(-1.6)} & 55.8 \textcolor{red}{(+2.0)} & 63.1 \textcolor{red}{(+2.3)} \\
             & CoT & 72.3 \textcolor{blue}{(-1.9)} & 53.6 \textcolor{red}{(+1.8)} & 63.7 \textcolor{red}{(+0.4)} & 55.9 \textcolor{red}{(+2.1)} & 61.4 \textcolor{red}{(+0.6)} \\
             \cmidrule{2-7}
             & ReAct & 81.9 \textcolor{red}{(+7.7)} & 51.6 \textcolor{blue}{(-0.2)} & 60.4 \textcolor{blue}{(-2.9)} & 52.9 \textcolor{blue}{(-0.9)} & 61.7 \textcolor{red}{(+0.9)} \\
             & + \texttt{Qwen72B} Assisted & 83.1 \textcolor{red}{(+8.9)} & 52.5 \textcolor{red}{(+0.7)} & 62.2 \textcolor{blue}{(-1.1)} & 54.5 \textcolor{red}{(+0.7)} & 63.1 \textcolor{red}{(+2.3)} \\ 
             \cmidrule{2-7}
             &\cellcolor[HTML]{ECF4FF} \name{} &\cellcolor[HTML]{ECF4FF} 83.8 \textcolor{red}{(+9.6)} &\cellcolor[HTML]{ECF4FF} 57.0 \textcolor{red}{(+5.2)} &\cellcolor[HTML]{ECF4FF} 63.2 \textcolor{blue}{(-0.1)} &\cellcolor[HTML]{ECF4FF} 56.4 \textcolor{red}{(+2.6)} &\cellcolor[HTML]{ECF4FF} 64.6 \textcolor{red}{(+3.8)}\\
             &\cellcolor[HTML]{ECF4FF}+ \texttt{Qwen72B} Assisted &\cellcolor[HTML]{ECF4FF} 92.7 \textcolor{red}{(+18.5)} &\cellcolor[HTML]{ECF4FF} 64.6 \textcolor{red}{(+12.8)} &\cellcolor[HTML]{ECF4FF} 64.0 \textcolor{red}{(+0.7)} &\cellcolor[HTML]{ECF4FF} 60.6 \textcolor{red}{(+6.8)} &\cellcolor[HTML]{ECF4FF} 70.5 \textcolor{red}{(+9.7)} \\ 
             &\cellcolor[HTML]{ECF4FF}+ \texttt{Qwen3} Assisted &\cellcolor[HTML]{ECF4FF} 90.7 \textcolor{red}{(+16.5)} &\cellcolor[HTML]{ECF4FF} 66.2 \textcolor{red}{(+14.4)} &\cellcolor[HTML]{ECF4FF} 67.2 \textcolor{red}{(+3.9)} &\cellcolor[HTML]{ECF4FF} 59.8 \textcolor{red}{(+6.0)} &\cellcolor[HTML]{ECF4FF} 71.0 \textcolor{red}{(+10.2)} \\ 
        \bottomrule
        \end{tabular}
        }
    \end{table*}

\subsection{Main Results} \label{sec: Results}

\paragraph{\name{} exhibits significant and consistent performance enhancement over baselines across all the benchmarks.} 
As listed in Table~\ref{table:main}, despite better performance than the direct method on MME dataset, VDGD and CCoT fail to demonstrate consistent improvements on the other datasets.
In contrast, \name{} consistently surpasses all other baselines across all benchmarks for every base model, enhancing the average performance of \texttt{GPT-4o-mini} by 7.9\%, demonstrating the superiority and task robustness of \name{}. 
Section~\ref{sec: more dataset} and Table~\ref{table: 4o addition} evaluate \name{} on additional benchmarks, including reasoning and VQA tasks, further proving its strong performance and effectiveness across various multi-modal tasks.

\paragraph{Decoupling the visual perception and textual reasoning capabilities of an LVLM outperforms their simultaneous inherent usage.} 
Table~\ref{table:main} illustrates that CoT, utilizing both capabilities simultaneously, does not exhibit consistent performance enhancements over the "Direct" method when applied to the \texttt{Llama3-LLaVA-NeXT-8B} and \texttt{Qwen2.5-VL-7B-Instruct} models.
In contrast, despite the same models, \name{} alternates between visual information acquisition and textual reasoning processes, allowing to leverage each capability more effectively. This enables \name{} to consistently outperform CoT with both \texttt{Llama3-LLaVA-NeXT-8B} and \texttt{GPT-4o-mini} across all benchmarks, validating the effectiveness of capability decoupling.

\paragraph{Proactive information acquisition surpasses peer passive methods}, especially in complex visual reasoning tasks.
Specifically, compared to MME, MathVista and HallusionBench present higher image complexity and question difficulty, and thus require stronger visual understanding and textual reasoning capabilities.
This leads to performance degradation of passive methods (\textit{i.e.}, VDGD and CCoT), highlighting their limited applicability to complex visual reasoning tasks.
In contrast, \name{} achieves notable performance improvements, up to 5.1\%, by proactively acquiring visual information from images rather than generating question-agnostic captions. This aligns with out previous obervations in Section~\ref{sec: passive analysis} that passive methods introduce substantial information redundancy or omission, misleading subsequent reasoning processes.

\paragraph{Text-only LLMs can be effectively integrated into \name{} for dramatically enhanced performance.}
As mentioned in Section~\ref{sec:advantages}, the decoupled visual perception and textual reasoning capabilities facilitate the seamless integration of text-only LLMs. 
To demonstrate the utility of this advantage, we fix the Vision Expert and replace the textual sub-agents in \name{} with text-only LLMs, according to the configuration in Table~\ref{table:text agent set}.
As listed in Table~\ref{table:main}, with the assistance of powerful existing LLMs, the \texttt{Llama3-LLaVA-NeXT-8B} Vision Expert receives remarkable performance boost across all benchmarks, particularly by 15\% on MMMU and 11.7\% on MathVista, compared to directly providing answers.
In addition, by configuring the Summarizer to \texttt{Qwen3-32B}, \name{} acquires the same kind of deep reasoning capability as \texttt{\texttt{Deekseek-R1}} and \texttt{OpenAI o1}, resulting in a 10.2\% average performance improvement for the \texttt{Qwen2.5-VL-7B-Instruct} Vision Expert.
In contrast, ReAct gains a much smaller improvement. This highlights the unique advantage of \name{} in leveraging existing text-only LLMs for enhanced performance. Notably, this advantage may open new avenues for continuously pushing the performance limits of LVLMs with the assistance of existing powerful LLMs.

\begin{table*}[!t]
\caption{Performance of different models across two visual benchmarks. Based on the performance of the base model(\textit{i.e.}, \texttt{Qwen2.5-VL-7B-Instruct}), red and blue signify the improvement and degradation, respectively.}
\label{table: proreason vl}
\centering
\begin{tabular}{c c c c c}
    \toprule
    \multirow{2}{*}{Dataset} & \multicolumn{4}{c}{Model} \\
    \cmidrule{2-5}& \texttt{Qwen2.5-VL-7B-Instruct} & \texttt{R1-Onevision-7B} & \texttt{\name{}-VL} & \texttt{\name{}-Q3}\\
    \midrule
    \multirow{-1}{*}{MME} 
     & 72.3& 88.1 \textcolor{red}{(+15.8)}& 91.2 \textcolor{red}{(+18.9)}& 90.8 \textcolor{red}{(+18.5)}\\[-0.2em] 
    \multirow{-1}{*}{MMMU(val.)} 
     & 53.6& 53.1 \textcolor{blue}{(-0.5)}& 65.4 \textcolor{red}{(+11.8)}& 67.4 \textcolor{red}{(+13.8)}\\[-0.2em] 
    \midrule
    \multirow{-1}{*}{Average} 
     & 63.0& 70.6 \textcolor{red}{(+7.6)} & 78.3 \textcolor{red}{(+15.3)} & 79.1 \textcolor{red}{(+16.1)}\\[-0.2em] 
    \bottomrule
\end{tabular}
\end{table*}

\begin{table*}[!t]
\caption{Average token and time consumption of multiple approaches with \texttt{GPT-4o-mini} model on the MME and MathVista benchmarks.}
\label{table: cost}
\centering
\begin{tabular}{c c c c c c c c}
    \toprule
         \multirow{2}{*}{Dataset} & \multirow{2}{*}{Method} & \multicolumn{6}{c}{\texttt{GPT-4o-mini}} \\
         \cmidrule{3-8}& & Direct & VDGD & CCoT & CoT & ReAct & \name{}\\
    \midrule
        \multirow{3}{*}{MME} 
         & Input & 393.9 & 1020.9 & 1024.3 & 403.9 & 1645.0 & 1286.8\\
         & Output & 5.9 & 155.6 & 254.3 & 103.4 & 197.0 & 327.2\\
         & Time(s) & 6.1 & 12.3 & 12.8 & 7.3 & 18.9 & 18.4\\
        \cmidrule{2-8}
         \multirow{3}{*}{MathVista}
         & Input & 368.4 & 955.5 & 961.2 & 375.4 & 3092.8 & 2238.6\\
         & Output & 51.8 & 263.3 & 307.3 & 479.7 & 845.1 & 788.6\\
         & Time(s) & 4.2 & 11.5 & 12.3 & 12.0 & 28.8 & 24.8\\
    \bottomrule
    \end{tabular}
\end{table*}

\subsection{Downstream Task Model Enhancement} \label{Sec: Model Enhancement}
\paragraph{High-quality visual reasoning data produced by \name{} significantly improves the performance of downstream task models.}
We select the test set that has no answer in the MMMU dataset and gather reasoning processes paired with corresponding answers on this set, generated by two configurations detailed in Table~\ref{table:text agent set}: \name{} + \texttt{Qwen72B} Assisted and \name{} + \texttt{Qwen3} Assisted.
After eliminating samples with inconsistent answers between these two configurations, we obtain two filtered datasets each containing 5,980 entries. 
These datasets are then used to fine-tune \texttt{Qwen2.5-VL-7B-Instruct} (Section~\ref{sec: Model Training}), yielding two models: \texttt{\name{}-VL} and \texttt{\name{}-Q3}. Notably, \texttt{\name{}-Q3} inherits the deep reasoning capabilities of \texttt{Qwen3-32B}, which incorporates a reasoning process containing a \texttt{<think>... </think>} part. 

Table~\ref{table: proreason vl} illustrates that while \texttt{R1-Onevision-7B} features deep reasoning capabilities and shows performance enhancements on MME, it fails to demonstrate improvement on the MMMU validation set. In contrast, both \texttt{\name{}-VL} and \texttt{\name{}-Q3} achieve performance gains exceeding 10\% on the MMMU validation set and also exhibit significant improvements on MME. 
This indicates that \name{} effectively transfers the exceptional reasoning capabilities of LLM-integrated decoupled systems to downstream task models, highlighting the potential for leveraging the robust abilities of existing LLMs to continuously enhance the performance of LVLMs.

\subsection{Efficiency and Complexity Analysis}
\label{sec:cost}
\paragraph{\name{} improves both efficiency and performance compared to the baseline method}. 
Table~\ref{table: cost} presents the evaluation of average token consumption and time expenditure of different methods on the MME and MathVista datasets. 
Notably, \name{} requires significantly fewer tokens than ReAct, while achieving superior performance over ReAct as analyzed in Section~\ref{sec: Results}. Moreover, compared to the visual reasoning frameworks VDGD and CCoT, which also involve multiple image inputs, \name{}'s token consumption on MME is only about 20\% higher, yet it achieves a 12.7\% performance improvement.
Regarding time efficiency, \name{} achieves 11.1\% higher performance than VDGD and CCoT on MME with comparable time consumption (18.4s/sample vs. VDGD's 12.3s and CCoT's 12.8s), while ReAct is slower than \name{} across datasets, highlighting \name{}'s efficiency advantage. 
Furthermore, as discussed in Section~\ref{Sec: Model Enhancement}, \name{}'s robust visual reasoning capabilities can be transferred to downstream task models, further ensuring the performance and efficiency of our approach.

\paragraph{\name{} achieves a balance between performance and efficiency through its Memory component and adaptive mechanisms}. The Memory component enables \name{} to keep compact information representations, avoiding lengthy reasoning traces like ReAct, reducing token use and boosting efficiency.
Additionally, \name{} dynamically adjusts Proactive Visual Perception iterations based on question difficulty, minimizing overhead for simple tasks while enhancing complex problem-solving. As shown in Table~\ref{table:iterations}, MathVista's greater challenge prompts more iterations compared to MME, resulting in higher token usage and longer reasoning times. This aligns with the difficulty levels of the datasets, demonstrating the adaptive nature of \name{}.

\subsection{Ablation Study and Further Analysis} \label{sec: more experiment}
\paragraph{Relative Importance of Sub-agents.}
To evaluate the importance of each sub-agent in \name{}, in Section~\ref{sec: The Importance of Different Sub-agents}, we replace Dispatcher, Vision Expert, Insight Expert, Referee, and Summarizer individually with the less capable \texttt{Llama3-LLaVA-NeXT-8B}(which demonstrates weaker visual understanding and textual reasoning capabilities), while keeping the other sub-agents as \texttt{GPT-4o-mini}. The performance degradation on the MME and MMMU benchmarks is then used to measure the significance of each sub-agent. Results indicate that the Summarizer is the most critical sub-agent, closely followed by Referee.

\paragraph{Which One is More Crucial: Visual Understanding or Textual Reasoning}
In Section~\ref{sec: Which is More Crucial}, we perform comparative experiments by substituting the vision expert and text sub-agents in the \name{} with \texttt{Llama3-LLaVA-NeXT-8B} and \texttt{GPT-4o-mini} respectively. Our findings indicate that while both visual understanding and textual reasoning capabilities are essential for multimodal tasks, textual reasoning ability holds greater significance in visual reasoning tasks. This result is consistent with our earlier analysis in Section~\ref{sec: The Importance of Different Sub-agents}, which identifies the Summarizer and Referee as the most critical sub-agents.

\paragraph{The Critical Implication of Decoupling.}
In Section~\ref{sec: Role of Decoupling}, to validate the necessity of decoupling visual perception and textual reasoning in \name{}, we systematically merge sub-agents through three configurations while preserving identical prompts and procedures, with observed performance degradation quantitatively demonstrating the critical role of decomposed processing in enhancing capabilities. The experimental results indicate that Decoupling serves as a crucial mechanism for improving \name{}'s performance in complex visual reasoning tasks

\paragraph{Reasoning Process Evaluation of \name{}.}
In Section~\ref{sec: llm evaluate}, we evaluate the responses generated by \name{} using LLMs. The analysis reveals that, compared to CoT, \name{} produces more relevant answers with reduced redundancy and deficiency, consistent with its enhanced performance.

\paragraph{Referee’s Dispel of Hallucinations.}
In Section \ref{Referee's Dispel of Hallucinations}, adhering to the settings in Section \ref{sec: implementation Details}, we assess \name{} (powered by \texttt{GPT-4o-mini}) on MMMU and HallusionBench with different attempt allowances (1/3/5). When attempts are unsuccessful, systematic Memory clearance is triggered (determined by the Referee's lack of sufficient information). As the number of attempts increases, the Referee has more opportunities for information filtering. The observed performance variations illustrate the crucial influence of the Referee's decision - making and filtering efficiency on the system's capabilities.
Experimental results indicate that the Referee module effectively filters hallucinated information to improve the visual comprehension capabilities of our framework.

\paragraph{Frequency of selection of various experts.}
In Section~\ref{Frequency of selection of various experts}, we assess how often the Dispatcher selects the Vision Expert or Insight Expert across both MME and MMMU benchmarks. The experimental results demonstrate that \name{} adaptively adjusts the frequencies of expert selection, leading to consistent performance improvements.

\section{Conclusion}
In this paper, we first validate that existing multi-modal reasoning approaches still suffer insufficient and irrelevant visual descriptions, as well as limited multi-modal capacities.
To address these issues, we decompose the visual reasoning process into visual perception and textual reasoning stages, and introduce a novel visual reasoning framework named \name{}, featuring decoupled vision-reasoning capabilities and multi-run proactive perception.
Empirically, extensive experiments demonstrate the superiority of \name{} over both passive image information acquisition methods and multi-step reasoning frameworks for text-only LLMs across multiple visual reasoning benchmarks with both open-source and closed-source models.
Notably, our method showcases the remarkable feasibility of integrating LLMs for multi-modal reasoning with dramatically improved performance, highlighting the great potential for LLM-assisted LVLM reasoning in future research.

\section{Limitations}

In this section, we analyze the limitations of the proposed method based on typical errors made by \texttt{GPT-4o-mini}-driven \name{}, as exemplified in Figures~\ref{fig:error case1} and \ref{fig:error case2}, to gain further understanding and identify potential research directions.

\paragraph{Cumulative Errors.}
As illustrated in Figure~\ref{fig:error case1}, the vision expert mistakenly perceives the clock as 6:25, which misguides the reasoning of subsequent agents and ultimately leads to an incorrect conclusion. More broadly, similar misperceptions occur frequently in errors made by \name{}. This indicates that, with the assistance of LLMs, \name{} has effectively addressed the reasoning deficiencies in multi-modal tasks, while the vision expert plays a significant role for further improvement of multi-modal capabilities.

\paragraph{Contradictory Information among Agents.}
Considering that multiple agents are engaged in the answering process, we try to find instances where contradictory information is provided by different agents, especially the vision expert and Insight Expert. However, as shown in Figure~\ref{fig:error case1} and \ref{fig:error case2}, when one agent (\textit{e.g.}, the vision expert) makes an error and the referee even hints at a possible mistake, other agents (\textit{e.g.}, the Insight Expert) tend to adhere to the available information instead of questioning it. This tendency results in a failure to find cases with contradictory information, and also highlights the importance of a reflection mechanism~\citep{ji2023towards} in agent collaboration, which is left for future exploration.

\paragraph{We also examined existing multi-step reasoning frameworks,} such as ReAct~\citep{yao2022react}, ToT~\citep{yao2024tree}, and Insight-V~\citep{dong2024insight}, aiming to find solutions for resolving accumulated errors and contradictory information. However, we found that these approaches may also fail to identify effective solutions and \textbf{do not discuss the impact of the aforementioned drawbacks}. We sincerely WELCOME any constructive discussions regarding accumulated errors and contradictory information!
\textbf{Besides, ProReason demonstrates significantly improved overall performance, suggesting fewer errors made by ProReason in the whole task distribution.}

\section{Ethics Statement}
We adhere strictly to the ACL Code of Ethics throughout our research. To our knowledge, the methods we introduce pose no foreseeable risks. We provide comprehensive details of the computing infrastructure used for all computational experiments in the paper, along with transparent statistics on our results and a detailed configuration of our experimental setup, including the optimal hyperparameter values. Furthermore, we will release the code upon publication to facilitate easy public reproducibility.

\section{Acknowledgements}
We want to thank our anonymous AC and reviewers for their feedback. This work was supported in part by Hong Kong Innovation and Technology Commission’s Innovation and Technology Fund (Award No. ITS/269/22FP), Hong Kong RGC grants  C7004-22G (CRF) and CRS\_PolyU501/23 (CRS).

\newpage
\bibliography{acl_latex}

\newpage
\appendix
\section{Appendix}\label{sec:appendix}

\subsection{Related work} \label{sec: related}
\paragraph{Large Visual-Language Model.} \label{sec: Large Visual-Language Models}
Recently, large vision-language models (LVLMs)~\citep{qwen2.5, QwenVlA_Bai2023,chen2023minigpt, liu2024visual} have 
garnered widespread attention and demonstrated remarkable advancements in understanding and generating multi-modal contents.
In the open-source domain, numerous LVLMs, like LLaVA~\citep{liu2023llava, liu2023improvedllava, liu2024llavanext, li2024llavanext-strong, li2024llava} and InternVL~\citep{chen2024internvl} families, have been extensively developed.
In the closed-source domain, proprietary models such as GPT-4o\footnote{\url{https://openai.com/index/hello-gpt-4o/}} and Gemini Pro 2.5\footnote{\url{https://deepmind.google/technologies/gemini/pro/}} have also achieved significantly success.
Additionally, multi-agent frameworks like VipAct~\citep{zhang2024vipact} have been developed to improve LVLMs' perception of visual details.
Despite these advancements, existing LVLMs still encounter challenges in effectively integrating visual understanding with textual reasoning capabilities simultaneously. This limitation is particularly evident in their diminished attention to image content during visual reasoning process, such as chart interpretation and visual math reasoning, leading to degraded performance~\citep{liu2023hallusionbench, ghosh2024vdgd} and motivating more effective solutions.

\paragraph{Multi-step Reasoning Framework.} \label{sec: Multi-step Reasoning Frameworks}
Multi-step reasoning frameworks improve LLM performance by breaking down complex tasks. Chain-of-Thought (CoT)~\citep{wei2022chain} enhances reasoning via explicit intermediate steps, demonstrating effectiveness in both textual and visual tasks~\citep{zhang2024improve}, while Tree-of-Thoughts (ToT)~\citep{yao2024tree} extends this by evaluating multiple reasoning paths. ReAct~\citep{yao2022react} integrates dynamic knowledge retrieval during reasoning.
In visual reasoning, several frameworks assist LVLM by extracting image information into text.
Visual Description Grounded Decoding (VDGD)~\citep{ghosh2024vdgd} describes the image and appends this description to the prompt, aiding LVLMs in visual reasoning tasks. Compositional Chain-of-Thought (CCoT)~\citep{mitra2024compositional} guides LVLMs to create scene graphs (SGs) that link visual and textual domains, supporting subsequent tasks. However, these methods often use a question-agnostic, reasoning-free visual extraction process, resulting in irrelevant or redundant information. 
Insight-V~\citep{dong2024insight} trains LVLM with a multi-agent system, but all agents rely on a single LVLM, blending visual understanding with textual reasoning and failing to address the challenge of effectively managing both. Unfortunately, this method does not release its code and prompts, resulting in low reproducibility and making comparison difficult.
In response to these drawbacks, we introduce \name{}, which decouples visual reasoning tasks into proactive visual perception (i.e., eyesight) and textual reasoning (i.e., wisdom), and makes all prompts available. By leveraging the strengths of the decoupled system, \name{} effectively integrates existing powerful LLMs to achieve high-performance visual reasoning and successfully transfers this capability to downstream task models.

\subsection{Dataset} \label{sec: Dataset}
To thoroughly assess the performance of our framework, we have carried out experimental evaluations using four benchmark datasets: the Multi-modal Large Language Model Evaluation (MME)~\citep{yin2023survey}, the cross-disciplinary Massive Multi-modal Understanding and Reasoning benchmark (MMMU)~\citep{yue2023mmmu}, the visual mathematical reasoning assessment MathVista~\citep{wang2024measuring}, and the multimodal illusion detection benchmark HallusionBench~\citep{liu2023hallusionbench}. Each of these benchmarks necessitates strong visual reasoning capabilities for successful task completion, and we provide concise descriptions below:
\begin{itemize}
    \item[\textbullet] \textbf{MME} is an inclusive benchmark that encompasses 14 subtasks, designed to evaluate perceptual and cognitive abilities. Given our focus on visual reasoning, we select the cognition-relevant tasks, including Commonsense Reasoning, Numerical Calculation, Text Translation, and Code Reasoning.

    \item[\textbullet] \textbf{MMMU} evaluates multi-modal models with multidisciplinary tasks that require college-level domain-specific knowledge and detailed reasoning. It comprises 11,500 questions across 30 disciplines and 183 sub-fields, emphasizing advanced perception and domain-specific reasoning.

    \item[\textbullet] \textbf{MathVista} focuses on more challenging mathematical reasoning tasks that demand precise visual recognition and compositional reasoning. It includes 6,141 examples from 31 multi-modal mathematics datasets.

    \item[\textbullet] \textbf{HallusionBench} 
    evaluates models' ability to reason with images such as statistical charts, emphasizing nuanced visual understanding. It consists of 346 images paired with 1,129 questions, meticulously crafted by experts. 
\end{itemize}

\subsection{Baselines} \label{sec: baseline_app}
\begin{enumerate}
\item[\textbullet] \textbf{Direct.} 
As indicated by the name, models are required to answer questions directly without dedicated prompts. This baseline is set to evaluate the initial performance of base models.
\item[\textbullet] \textbf{CoT.} CoT is an advanced prompting method that encourages LLMs to break complex tasks down into a series of easy steps, which has been applied broadly and verified to boost the reasoning performance remarkably~\citep{cot_survey}. 
\item[\textbullet] \textbf{ReAct.} ReAct is an LLM-specific agent framework, which performs tasks by alternating between reasoning and execution behaviors. To extend it to multi-modal domain, we use two LVLMs to perform both steps, and rename them as the Vision and Insight Experts, respectively. This aligns with our notions for easy understanding, and is shown in Figure~\ref{fig:method}. 
\item[\textbullet] \textbf{VDGD.} 
VDGD involves two main steps: initially, LVLMs generate detailed image captions, which are then incorporated into prompts to aid inference. During the inference process, VDGD also utilizes a formula based on Kullback-Leibler divergence to select tokens that minimally deviate from the description, thereby enhancing the relevance of the model's reasoning to the image\footnote{Since we cannot obtain the tokens output by \texttt{GPT-4o-mini}, we omit the step of selecting the token with the smallest deviation from the image description when implementing VDGD for \texttt{GPT-4o-mini}.}.
\item[\textbullet] \textbf{CCoT.}  
Given an image and the question, CCOT first generates a scene graph of the image with LVLMs, and then extracts the answer by prompting the LVLMs with the graph. 
\item[\textbullet] \textbf{R1-Onevision.} 
\texttt{R1-Onevision-7B} is an advanced multimodal reasoning model based on \texttt{Qwen2.5-VL-7B-Instruct} that has deep thinking capabilities akin to Deepseek-R1 by transforming images into structured textual representations and employing a training framework that merges supervised fine-tuning with reinforcement learning.

\end{enumerate}

\subsection{Model Training} \label{sec: Model Training}
we use Supervised Fine-Tuning(SFT) and employ the parameter-efficient fine-tuning method \textbf{LoRA}~\citep{hu2021loralowrankadaptationlarge}. Specifically, we uniformly set the learning rate to \(1\times10^{-4}\), \(\texttt{lora\_dropout}=0\), and train the \texttt{\name{}-VL} for 1 epoch and \texttt{\name{}-Q3} for 2 epoch. These parameters are the optimal values obtained after multiple attempts.

\subsection{Supplementary Results and Analysis}\label{sec: Supplementary Results and Analysis}
\begin{table}[!h]
\centering
\caption{Performance of three recent LVLMs on MMMU dataset with different assisting techniques.}
\label{table:motivation}
\resizebox{\linewidth}{!}{
    \begin{tabular}{cccc}
        \toprule
            \multirow{2}{*}{Model} & \multicolumn{3}{c}{Method} \\
        \cmidrule{2-4}
            & Direct & CoT & VDGD \\
        \midrule
            \texttt{Llama3-LLaVA-NeXT-8B} & 41.8 & 41.5 & 42.7 \\
            \texttt{Qwen2.5-VL-7B-Instruct} & 51.8 & 52.2 & 52.6 \\
        \bottomrule
    \end{tabular}
}
\end{table}

\begin{table}[!h]
\centering
\caption{Effectiveness evaluation of passive captions along Detail Level, Question Relevance, and Reasoning Effective Info Inclusion. ``True'' and ``False'' denote the response correctness of \texttt{Llama3-LLaVA-NeXT-8B}.}
\label{table:cap_evalue}
\resizebox{\linewidth}{!}{
\begin{tabular}{ccc}
    \toprule
        \multirow{2}{*}{Score} & \multicolumn{2}{c}{\makebox[4cm]{\texttt{Llama3-LLaVA-NeXT-8B}}} \\
        \cmidrule{2-3}
        & \makebox[2cm]{True} & \makebox[2cm]{False} \\
    \midrule
        Detail Level   & 4.43 & 3.93 \\
        Question Relevance & 3.87 & 3.30 \\
        Reasoning Effective Info Inclusion & 3.91 & 3.57 \\
    \bottomrule
\end{tabular}
}
\end{table}

\subsubsection{Relative Importance of Sub-agents} \label{sec: The Importance of Different Sub-agents}
To assess the importance of each sub-agent within the \name{} framework for visual reasoning tasks, we design five scenarios where \texttt{Llama3-LLaVA-NeXT-8B} acts as Dispatcher, Vision Expert, Insight Expert, Referee, or Summarizer, respectively, while the other sub-agents are powered by \texttt{GPT-4o-mini}.
Given that \texttt{Llama3-LaVA-NeXT-8B} exhibits weaker visual understanding and textual reasoning capabilities than \texttt{GPT-4o-mini}, the more significant the performance drop incurred by replacing a sub-agent with \texttt{Llama3-LaVA-NeXT-8B} is, the more important that sub-agent is. Here we primarily consider the MME and MMMU benchmarks due to their comprehensive question coverage. The experimental results are presented in Table~\ref{table:sub-agent}.

\paragraph{Summarizer is the most crucial sub-agent, closely followed by Referee.} The replacement of Summarizer results in the most notable performance decline on both MME and MMMU tasks, reaching 6.2\% and 10.6\%, respectively. 
This highlights the critical function of the Summarizer in integrating all available information to conclude final answers. Besides, the substitution of Referee leads to a 10.1\% reduction on MMMU. Given that MMMU is more challenging than MME, this finding underscores the essential role of the Referee in assessing the sufficiency of information, particularly in more complex visual reasoning tasks. The analysis in Section~\ref{Referee's Dispel of Hallucinations} also demonstrates that the Referee plays a crucial role in enabling \name{} to accurately interpret visual detail information.

\paragraph{Relatively, Dispatcher and Insight Expert are the least essential sub-agents.}
Specifically, despite a decline, these two sub-agents exhibit significantly less performance degradation than other sub-agents. 
This can be attributed to the easier task of the Dispatcher, which requires minimal textual reasoning capabilities, and the infrequent calls of the Insight Expert, which is only activated when additional information needs to be inferred—a situation that is rare in current benchmarks.
Besides, both sub-agents operate within the acquisition loop, allowing for greater error tolerance. Even if some error occurs, subsequent iterations can compensate for the missing information.

In summary, each sub-agent contributes to the performance of \name{}, underscoring their necessity. Relatively, the Summarizer and Referee are the most critical sub-agents, while the Dispatcher and Insight Expert have the least impact.

\begin{table}[!h]
\caption{Performance of \name{} across five scenarios for sub-agent assessment on visual reasoning tasks. For each scenario, one sub-agent is replacing with \texttt{Llama3-LLaVA-NeXT-8B}, while the others are performed by \texttt{GPT-4o-mini}.
The blue text indicates the performance decline compared to the scenario with all agents performed by \texttt{GPT-4o-mini}.}
\label{table:sub-agent}
\centering
\resizebox{\linewidth}{!}{
\begin{tabular}{cccc}
\toprule
    \multirow{2}{*}{Model} & \multirow{2}{*}{Agent} & \multicolumn{2}{c}{Dataset} \\
\cmidrule{3-4}
        & & MME & MMMU \\
\midrule
    \multicolumn{2}{c}{\texttt{GPT-4o-mini}} & 90.4 & 61.6 \\
\cmidrule(rl){1-4}
    \multirow{5}{*}{\begin{tabular}{c} 
                        \texttt{Llama3-}\\ 
                        \texttt{LLaVA-} \\ 
                        \texttt{NeXT-8B} 
                    \end{tabular}} 
        & Dispatcher & 88.8 \textcolor{blue}{(-1.6)} & 60.9 \textcolor{blue}{(-0.7)} \\
        & Vision Expert & 84.7 \textcolor{blue}{(-5.7)} & 54.5 \textcolor{blue}{(-7.1)}\\
        & Insight Expert & 88.7 \textcolor{blue}{(-1.7)} & 60.2 \textcolor{blue}{(-1.4)}\\
        & Referee & 89.6 \textcolor{blue}{(-1.1)} & 51.5 \textcolor{blue}{(-10.1)} \\
        & Summarizer & 84.2 \textcolor{blue}{(-6.2)} & 51.0 \textcolor{blue}{(-10.6)}\\
\bottomrule
\end{tabular}
}
\end{table}

\subsubsection{Which One is More Crucial: Visual Understanding or Textual Reasoning?} \label{sec: Which is More Crucial}
\name{} effectively decouples the visual understanding and textual reasoning capabilities of LVLMs. However, it remains unclear which of these two capacities is more critical for visual reasoning tasks. To answer this question, we conduct comparative experiments of the following three scenarios:
\begin{enumerate}
\item[\textbullet]
\textbf{\texttt{Llama3-LLaVA-NeXT-8B} as All Sub-Agents}. All sub-agents within \name{} framework are performed by \texttt{Llama3-LLaVA-NeXT-8B} model.

\item[\textbullet]
\textbf{\texttt{GPT-4o-mini} as Vision Expert}. 
Based on the above scenario, we implement the Vision Expert with \texttt{GPT-4o-mini}, while keep the other textual sub-agents unchanged.

\item[\textbullet]
\textbf{\texttt{GPT-4o-mini} as Textual Sub-Agents.} 
Reversely, we utilize \texttt{Llama3-LLaVA-NeXT-} \texttt{8B} as the Vision Expert, and \texttt{GPT-4o-mini} for the other vision-irrelevant sub-agents.
\end{enumerate}

\paragraph{Textual reasoning capabilities outweigh visual understanding for multi-modal reasoning tasks, although both are important.} As shown in Table~\ref{table:improtance}, replacing either the Vision Expert or the other agents with the more capable \texttt{GPT-4o-mini} achieves consistent performance enhancement, highlighting the significance of both capabilities. However, substituting the textual sub-agents with \texttt{GPT-4o-mini} results in a more substantial performance boost compared to replacing the Vision Expert. This underscores the greater importance of textual reasoning over visual understanding for multimodal reasoning tasks, aligning with our previous analysis in Section~\ref{sec: The Importance of Different Sub-agents} that identifies the Summarizer and Referee as the most crucial sub-agents.

\begin{table}[!h]
\centering
\caption{Performance of \name{} with different configurations for the relative importance assessment between visual understanding and textual reasoning capabilities on visual reasoning tasks.
The red text highlights the performance improvements brought about by the introduction of \texttt{GPT-4o-mini}.
}
\label{table:improtance}
\resizebox{\linewidth}{!}{
\begin{tabular}{cccc}
\toprule
    \multirow{2}{*}{Model} & \multirow{2}{*}{Agent} & \multicolumn{2}{c}{Dataset} \\
\cmidrule{3-4}
        & & MME & MMMU \\
\midrule
    \multirow{2}{*}{\begin{tabular}{c} 
                            \texttt{GPT-4o} \\ 
                            \texttt{-mini} 
                        \end{tabular}}
    & Textual Sub-Agents &  84.7 \textcolor{red}{(+13.2)} & 54.5 \textcolor{red}{(+2.0)} \\
    & Vision Expert & 77.8 \textcolor{red}{(+6.3)} & 53.4 \textcolor{red}{(+0.9)} \\
\cmidrule(rl){1-4}
    \multirow{3}{*}{\begin{tabular}{c} 
                        \texttt{Llama3-}\\ 
                        \texttt{LLaVA-} \\ 
                        \texttt{NeXT-8B} 
                    \end{tabular}} 
    & All Sub-Agents & 71.5 & 52.5 \\
    & COT & 58.8 & 41.5 \\
    & Direct & 61.5 & 41.8 \\
\bottomrule
\end{tabular}
}
\end{table}

\subsubsection{The Critical Implication of Decoupling} \label{sec: Role of Decoupling}
To validate the critical implication of decoupling visual perception and textual reasoning in \name{} while eliminating potential confounding factors from prompt engineering and multiple CoT implementations, we systematically integrate sub-agents through three configurations:
\begin{enumerate}
\item[\textbullet]
Merge the Vision Expert and Insight Expert into a single sub-agent to examine the necessity of modality decoupling during sub-task execution.
\item[\textbullet]
Integrate the Dispatcher, Vision Expert, Insight Expert, and Referee as a unified agent to verify the essentiality of the process design for the Proactive Visual Perception stage.
\item[\textbullet]
Fully consolidate all five original sub-agents to demonstrate the pivotal role of decomposing visual reasoning tasks into distinct Proactive Visual Perception and Textual Reasoning phases.
\end{enumerate}

\begin{figure*}[t]
\centering
\includegraphics[page=3, width=\linewidth]{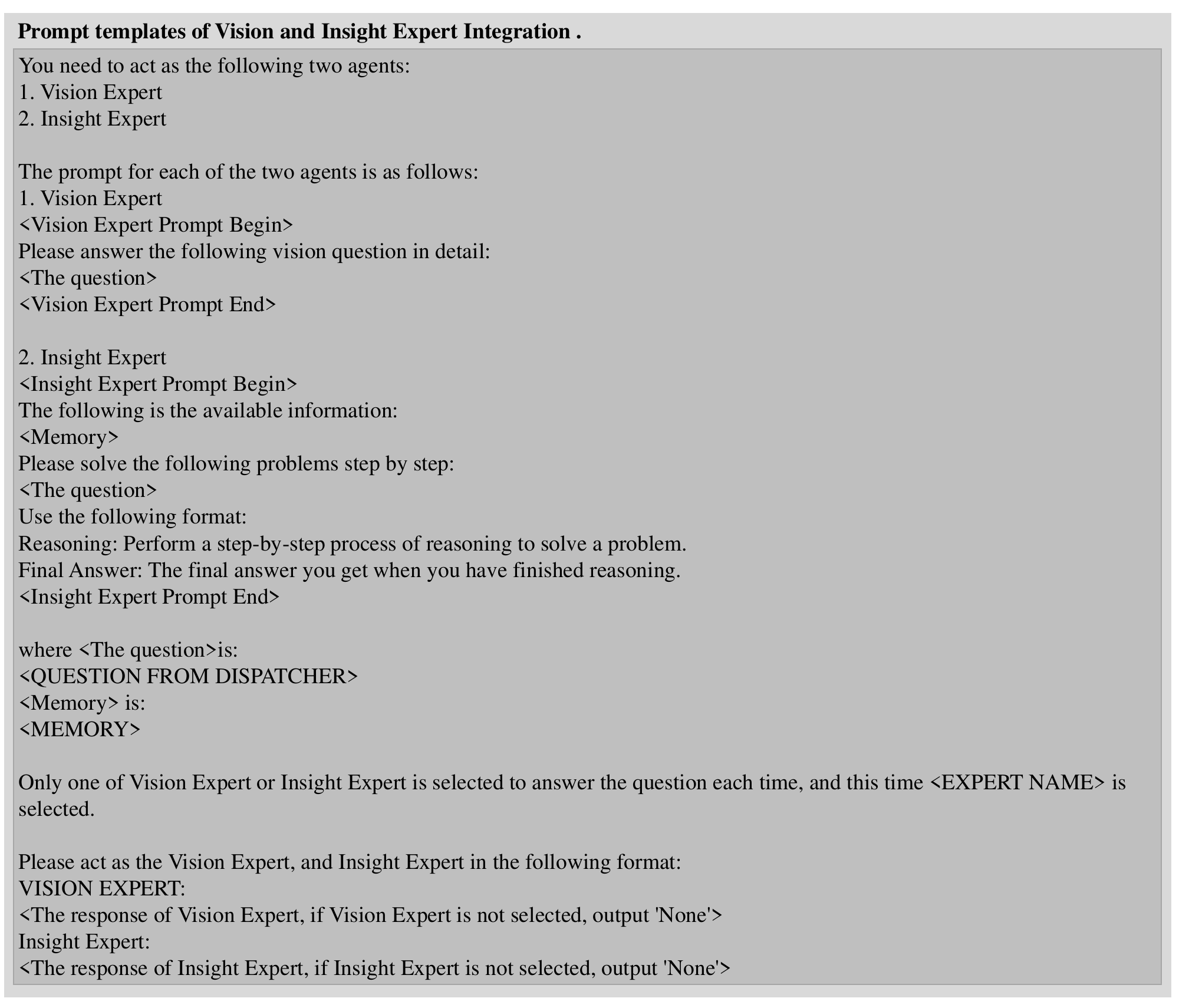}
\caption{
Prompt templates of Vision and Insight Expert Integration.
}
\label{fig:merge rv}
\end{figure*}

\begin{figure*}[t]
\centering
\includegraphics[page=2, width=\linewidth]{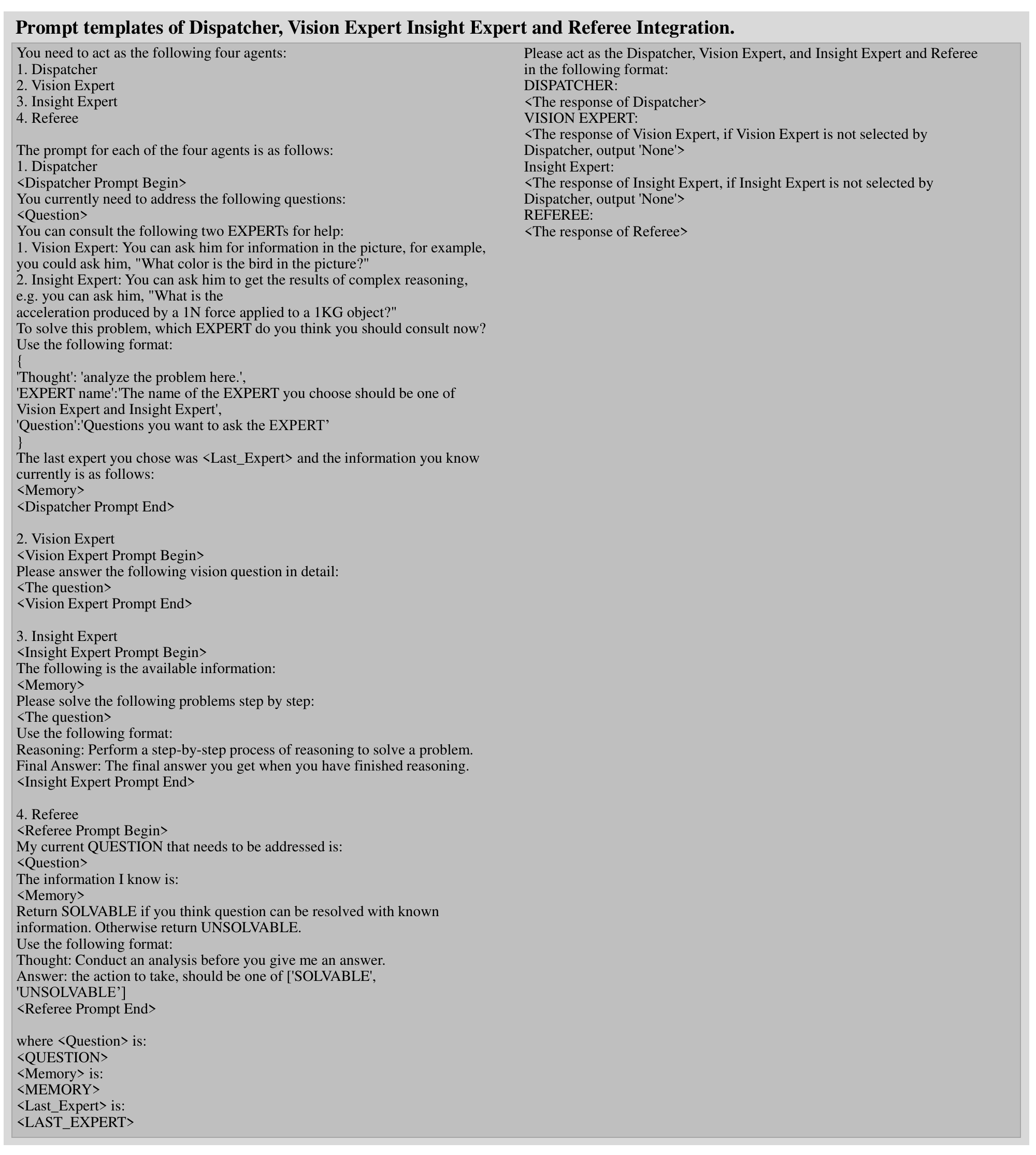}
\caption{
Prompt templates of Dispatcher, Vision Expert Insight Expert and Referee Integration.
}
\label{fig:merge dvrj}
\end{figure*}

\begin{figure*}[t]
\centering
\includegraphics[page=1, width=\linewidth]{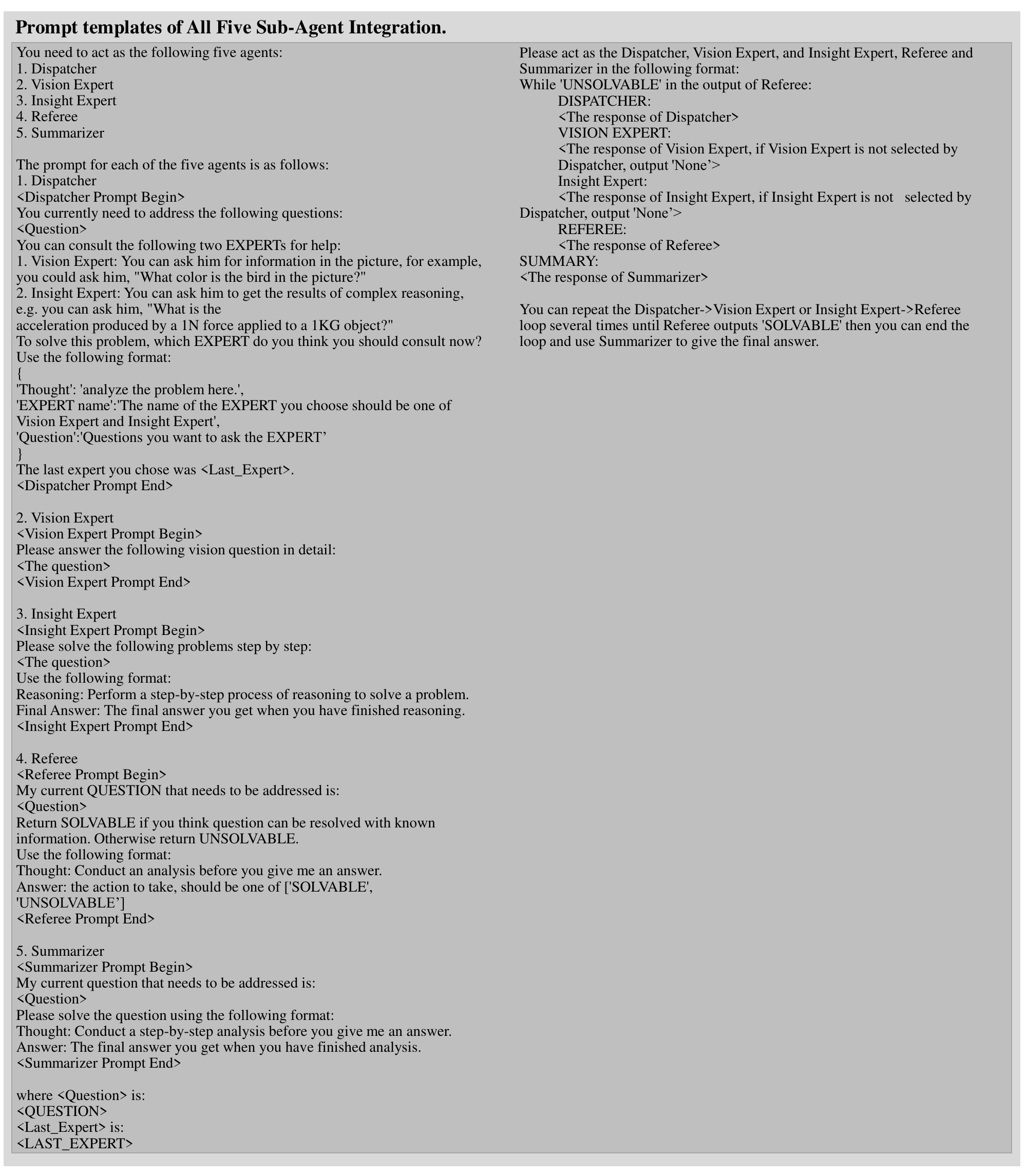}
\caption{
Prompt templates of All Five Sub-Agent Integration.
}
\label{fig:merge dvrjs}
\end{figure*}

The merged agents preserve identical prompts and maintain the same task execution procedures as their original counterparts, with their prompts shown in Figures~\ref{fig:merge rv}, \ref{fig:merge dvrj}, and \ref{fig:merge dvrjs}. Therefore, the performance degradation caused by agent merging quantitatively demonstrates the critical implication of decoupled processing in enhancing \name{}'s capabilities.

\paragraph{Decoupling serves as a crucial mechanism for improving \name{}'s performance in complex visual reasoning tasks}. As shown in Table~\ref{table:decoupling}, merging Vision and Insight Experts results in a 3\% performance drop on the MMMU benchmark, while combining Proactive Visual Perception with Textual Reasoning leads to a more significant 4.8\% reduction (56.8 vs. CoT's 58.5). Although agent merging also causes performance declines on MME, the merged versions still outperform CoT. Notably, given MMMU's substantially higher complexity compared to MME, these findings reveal that the decoupling of visual-textual processing fundamentally drives its performance gains in complex scenarios.

\begin{table}[!h]
\centering
\caption{
Impact of decoupling visual perception and textual reasoning on \name{} performance with results shown from sub-agent consolidation experiments.
The blue text highlights the performance degradation due to sub-agent integration.}
\label{table:decoupling}
\resizebox{\linewidth}{!}{
\begin{tabular}{ccc}
\toprule
    \multirow{2}{*}{\texttt{GPT-4o-mini}} & \multicolumn{2}{c}{Dataset} \\
\cmidrule{2-3}
        & MMMU & MME \\
\midrule
    \name{} & 61.6 & 91.9 \\
\midrule
    CoT & 58.5 & 87.8 \\
\cmidrule{1-3}
    \begin{tabular}{c}
         Vision \& Reasoning \\Expert
         Integration \\
    \end{tabular} & 58.6 \textcolor{blue}{(-3.0)} & 90.4 \textcolor{blue}{(-1.5)}\\
\cmidrule{1-3}
    \begin{tabular}{c}
        Dispatcher, \\ Vision \& Insight Expert,\\ \& Referee 
        Integration \\
    \end{tabular} & 57.8 \textcolor{blue}{(-3.8)} & 90.8 \textcolor{blue}{(-1.1)} \\
\cmidrule{1-3}
    All Sub-Agents Integration & 56.8 \textcolor{blue}{(-4.8)} & 89.2 \textcolor{blue}{(-2.7)} \\
\bottomrule
\end{tabular}
}
\end{table}

\subsubsection{Reasoning Process Evaluation of \name{}} \label{sec: llm evaluate}
\begin{figure*}[b]
\centering
\includegraphics[width=\linewidth]{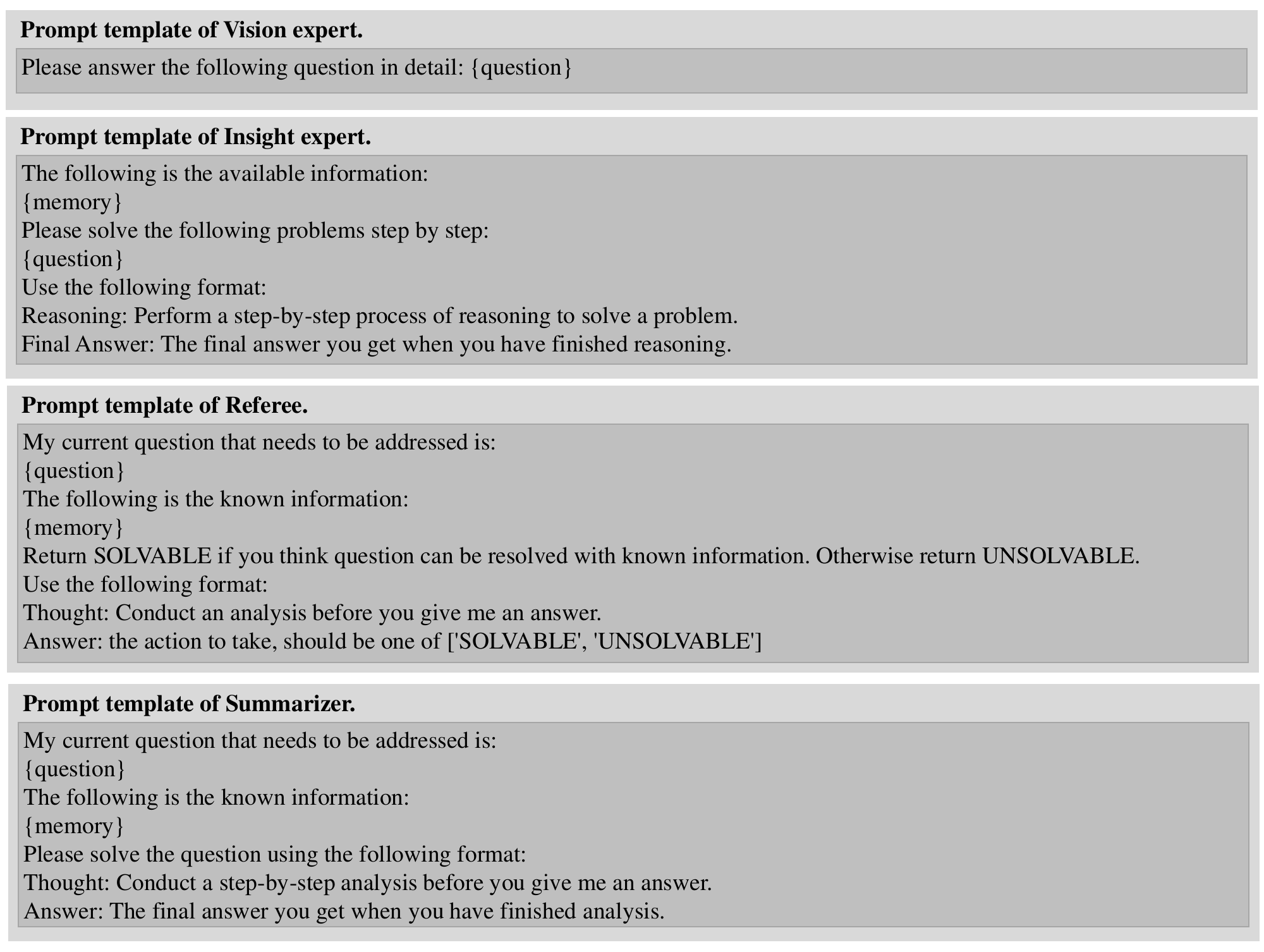}
\caption{Prompt templates of Vision Expert, Insight Expert, Referee, and Summarizer.}
\label{fig:prompt_else}


\centering
\includegraphics[width=\linewidth]{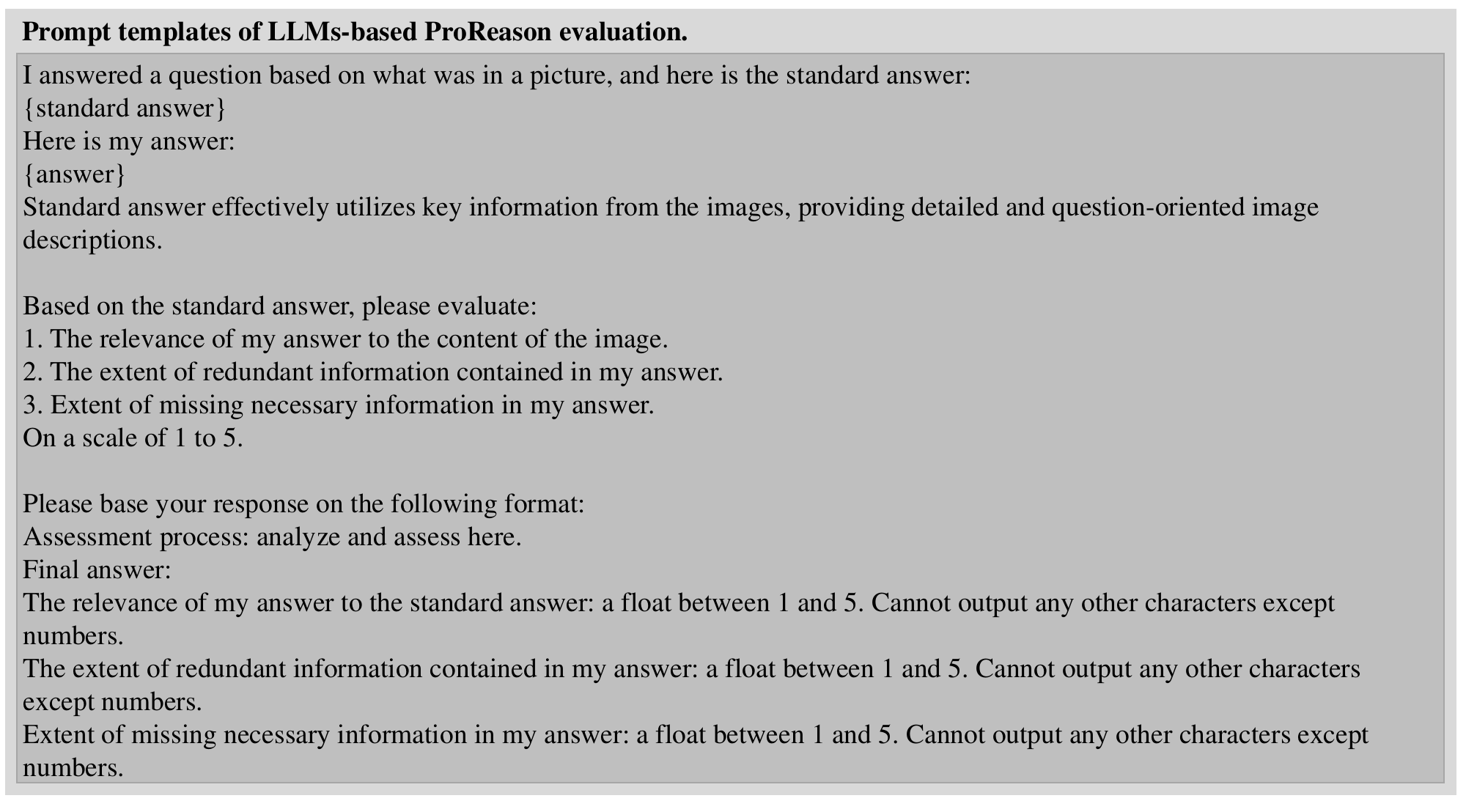}
\caption{Prompt templates of LLMs-based \name{} evaluation.}
\label{fig:prompt_ProReason_LLM}
\end{figure*}

Inspired by \citet{liu2023llava}, we design a pipeline using LLMs to further analyze the performance of \name{}. Given that \texttt{GPT-4o} achieved a 70.3\% score on MMMU, significantly surpassing \texttt{GPT-4o-mini}'s 58.5\% and \name{} driven by \texttt{GPT-4o-mini}'s 61.6\%, we adopt \texttt{GPT-4o}'s CoT-based answers on MMMU as the standard answer. Using this standard answer, we evaluate the reasoning process of \texttt{GPT-4o-mini} (CoT), the Memory of \name{} powered by \texttt{GPT-4o-mini} (\name{}-Memory), and the reasoning process of \name{}'s Summarizer (\name{}-Summarizer) on MMMU.
This assessment focuses on three key metrics: the relevance to standard answer (RE ↑), the degree of redundant information (RI ↓), and the extent of missing information (MI↓), where arrows indicate the directions of improvement.
The evaluation process is driven by \texttt{GPT-4o} and the prompt is shown in Figure~\ref{fig:prompt_ProReason_LLM}.

Specifically, compared to \texttt{GPT-4o-mini} (CoT), \textbf{\name{}-Summarizer produces more relevant answers with less redundancy and deficiency}, aligning with its improved performance. Compared to \name{}-Summarizer, \name{}-Memory exhibits the same RE, higher RI and lower MI scores. This suggests that \name{} allows some redundancy to prevent information loss in memory, as the former typically leads to more serious consequences than the latter. Subsequently, Summarizer can leverage its powerful reasoning capabilities to select the most relevant memory. 
\begin{table}[b]
\centering
\caption{Performance of \name{} driven by \texttt{GPT-4o-mini} assessed by LLMs compared to CoT on MMMU benchmark. Mainly includes three key metrics: the relevance to standard answers (RE ↑), the degree of redundant information (RI ↓), and the extent of missing information (MI ↓), where arrows indicate the directions of improvement.}
\label{table: llm evaluate proreason}
\resizebox{\linewidth}{!}{
\begin{tabular}{cccc}
\toprule
 \multirow{2}{*}{\texttt{GPT-4o-mini}} & \multicolumn{3}{c}{Metrics}\\
 \cmidrule{2-4}
    & RE ↑ & RI ↓ & MI ↓ \\ 
\midrule
     COT & 4.67 & 3.33 & 1.40 \\
     \name{}-Memory & 4.83 & 3.66 & 1.17 \\
     \name{}-Summarizer & 4.83 & 2.88 & 1.33 \\
\bottomrule
\end{tabular}}
\end{table}

\subsubsection{Referee's Dispel of Hallucinations} \label{Referee's Dispel of Hallucinations}
Following the implementation details outlined in Section \ref{sec: implementation Details}, we evaluate the \name{} system powered by \texttt{GPT-4o-mini} on both MMMU and HallusionBench under varying attempt allowances (\textit{i.e.}, 1, 3, and 5 attempts). Every unsuccessful attempt reflects the Referee's persistent determination that the information stored in Memory is insufficient to solve the problem. Before each new attempt, Memory is systematically cleared, ensuring the removal of information deemed irrelevant by the Referee's assessment. Consequently, increased attempt allowances essentially empower the Referee with enhanced opportunities for information filtration. The observed performance variations of \name{} across different attempt quotas demonstrate the critical impact of Referee's decision-making mechanism and information filtering efficacy on system capability.

\paragraph{The Referee module effectively filters hallucinated information to enhance the visual comprehension capabilities of our framework}. As demonstrated in Table~\ref{table:Referee's Dispel of Hallucinations}, the performance improvement on HallusionBench (2.8\%) significantly outpaces that on MMMU (1.5\%) as attempt opportunities increase from 1 to 5. Given HallusionBench's dual emphasis on reasoning proficiency and precise evaluation of visual details/hallucination control compared to MMMU, these results suggest that the Referee mechanism can effectively identify erroneous or irrelevant visual information, thereby strengthening \name{}'s capacity for meticulous visual understanding.

\begin{table}[!h]
\centering
\caption{Impact on HallusionBench and MMMU performance across different attempt allowances.}
\label{table:Referee's Dispel of Hallucinations}
\begin{tabular}{cccc}
\toprule
    \multirow{2}{*}{Dataset} & \multicolumn{3}{c}{Attempt Allowances} \\
\cmidrule{2-4}
     & 1 & 3  & 5 \\
\midrule
    MMMU & 60.1 & 59.9 & 61.6 \\
    HallusionBench & 57.1 & 58.9 & 59.9 \\
\bottomrule
\end{tabular}
\end{table}

\subsubsection{Frequency of selection of various experts} \label{Frequency of selection of various experts}
As listed in Table~\ref{table:frequency}, we evaluate the frequency of the Dispatcher choosing the Vision Expert or Insight Expert on both MME and MMMU benchmarks, with MMMU requiring higher visual and reasoning abilities. Specifically, compared to MME, the frequencies for both the Vision and Insight Experts are higher on the MMMU benchmark, aligning with their difficulty levels. Together with the results in Table~\ref{table:main} of our submission, \textbf{\name{} can adaptatively increase the frequencies of experts, and provide consistent performance improvements} (\textit{i.e.}, 11.2\% and 13.2\%). Despite the lower frequency of the Insight Expert, the significant performance enhancement highlights the importance of LLM-assisted reasoning capabilities for reasoning-essential questions. Additionally, the frequency of the Vision expert exceeding 1 underscores the importance of referees, which controls the loop to call experts multiple times, alleviating the issue of insufficient information.

\begin{table}[!h]
\centering
\caption{Frequency of the Dispatcher choosing the Vision Expert or Insight Expert on both MME and MMMU benchmarks.}
\label{table:frequency}
\resizebox{\linewidth}{!}{
\begin{tabular}{ccc}
\toprule
\multirow{2}{*}{Dataset} & \multicolumn{2}{c}{\texttt{GPT-4o-mini}} \\
\cmidrule{2-3}
     & Vision Expert & Insight Expert \\
\midrule
    MME & 1.16 & 0.12 \\
    MMMU & 1.64 & 0.38 \\
\bottomrule
\end{tabular}
}
\end{table}

\subsubsection{Performance Evaluation of \name{} on Supplementary Datasets} \label{sec: more dataset}
To further assess the effectiveness of \name{} across diverse visual tasks, we expand upon the four datasets introduced in Section~\ref{sec: setup} by incorporating three additional benchmarks: MathVerse~\citep{zhang2024mathverse}, MMStar~\citep{chen2024we}, and A-OKVQA~\citep{schwenk2022okvqa}. MathVerse focuses on visual reasoning within mathematical contexts, while MMStar highlights visual dependency and data reliability. \textbf{A-OKVQA, in contrast, serves as a knowledge-based VQA dataset requiring minimal reasoning.} For MathVerse, we utilize the Text Lite subset, which prioritizes visual reasoning by minimizing textual content. In MMStar, subsets related to coarse and fine-grained perception that are unrelated to reasoning are excluded. Table \ref{table: 4o addition} presents \name{}'s performance on each dataset, showcasing significant improvements across various forms of multi-modal tasks and \textbf{highlighting the effectiveness and generalizability of the perception and reasoning decoupling approach}.

\subsubsection{The Impact of Iterative Reasoning on \name{}'s Performance} \label{sec: iterative time}
As shown in the Table \ref{table: proreason_iterations}, we evaluated the performance of \name{}, powered by \texttt{GPT-4o-mini}, on the MME, MMMU, and MathVista datasets, given different maximum loop iterations. The performance of \name{} is observed to enhance, as the maximum iterations increase up to three, after which the performance tends to stabilize. This observation indicates that most questions in the MME, MMMU, and MathVista benchmarks can be effectively addressed within three cycles of question-oriented visual information extraction. \textbf{In cases of a few difficult questions, \name{} can adaptively extend the number of iterations}, proactively gathering more information from the image to arrive at the solution.

\begin{table}[!h]
\centering
\caption{Number of iterations required for each problem during the Proactive Visual Perception phase on MME and MathVista benchmarks.}
\label{table:iterations}
\resizebox{\linewidth}{!}{
\begin{tabular}{ccc}
\toprule
\multirow{2}{*}{Model} & \multicolumn{2}{c}{Dataset} \\
\cmidrule{2-3}
     & MME & MathVista \\
\midrule
    \texttt{GPT-4o-mini} & 1.28 & 2.13 \\
\bottomrule
\end{tabular}}
\end{table}

\begin{table*}[!t]
\caption{Performance of multiple approaches with \texttt{GPT-4o-mini} across 7 visual benchmarks. ``Hallu.'' is the abbreviation of HallusionBench. Based on the performance of the direct method, red and blue signify the improvement and degradation, respectively.}
\label{table: 4o addition}
\centering
\resizebox{\textwidth}{!}{
\begin{tabular}{c c c c c c c c c}
    \toprule
         \multirow{2}{*}{Model} & \multirow{2}{*}{Method} & \multicolumn{7}{c}{Dataset} \\
         \cmidrule{3-9}& & MME & MMMU & MathVista & Hallu. & MathVerse & MMStar & A-OKVQA \\
    \midrule
        \multirow{7}{*}{\texttt{GPT-4o-mini}} 
         & Direct & 79.2 & 48.4 & 53.0 & 56.0 & 28.2 & 46.2 & 78.6  \\
         & VDGD & 82.3 \textcolor{red}{(+3.1)} & 51.4 \textcolor{red}{(+3.0)} & 51.2 \textcolor{blue}{(-1.8)} & 52.4 \textcolor{blue}{(-3.6)} & 30.1 \textcolor{red}{(+1.9)} & 47.0 \textcolor{red}{(+0.8)} & 79.4 \textcolor{red}{(+0.8)} \\
         & CCoT & 80.8 \textcolor{red}{(+1.6)} & 54.2 \textcolor{red}{(+5.8)} & 53.6 \textcolor{red}{(+0.6)} & 56.7 \textcolor{red}{(+0.7)} & 29.2 \textcolor{red}{(+1.0)} & 45.4 \textcolor{blue}{(-0.8)} & 79.2 \textcolor{red}{(+0.6)}\\
         & CoT & 87.8 \textcolor{red}{(+8.6)} & 58.5 \textcolor{red}{(+10.1)} & 53.8 \textcolor{red}{(+0.8)} & 56.3 \textcolor{red}{(+0.3)} & 28.9 \textcolor{red}{(+0.7)} & 47.2 \textcolor{red}{(+1.0)} & 80.9 \textcolor{red}{(+2.3)}\\
         & ReAct & 87.3 \textcolor{red}{(+8.1)} & 54.8 \textcolor{red}{(+6.4)} & 49.3 \textcolor{blue}{(-3.7)} & 51.1 \textcolor{blue}{(-4.9)} & 30.4 \textcolor{red}{(+2.2)} & 46.7 \textcolor{red}{(+0.5)} & 80.6 \textcolor{red}{(+2.0)}\\
         & \name{} & 91.9 \textcolor{red}{(+12.7)} & 61.6 \textcolor{red}{(+13.2)} & 54.9 \textcolor{red}{(+1.9)} & 59.9 \textcolor{red}{(+3.9)} & 31.6 \textcolor{red}{(+3.4)} & 49.1 \textcolor{red}{(+2.9)} & 81.3 \textcolor{red}{(+2.7)}\\
         \cmidrule{2-9}
          & \textbf{Average} & 84.63 & 54.82 & 52.63 & 55.4 & 29.7 & 46.9 & 80.0   \\
    \bottomrule
    \end{tabular}
    }
\end{table*}

\begin{table*}[!t]
\caption{Performance of ProReason with \texttt{GPT-4o-mini} across MME, MMMU, and MathVista benchmarks, given different maximum loop iterations. The numbers in parentheses signify the improvement or degradation over the direct method, with red for improvement and blue for degradation.}
\label{table: proreason_iterations}
\centering
\resizebox{\textwidth}{!}{
\begin{tabular}{c c c c c c c c c c c}
    \toprule
    \multirow{2}{*}{Dataset} & \multirow{2}{*}{Direct} & \multirow{2}{*}{VDGD} & \multirow{2}{*}{CCoT} & \multirow{2}{*}{CoT} & \multirow{2}{*}{ReAct} & \multicolumn{5}{c}{ProReason (max loop iterations)} \\
    \cmidrule{7-11}
    & & & & & & 1 & 2 & 3 & 4 & 5 \\
    \midrule
    MME & 79.2 & 82.3 \textcolor{red}{(+3.1)} & 80.8 \textcolor{red}{(+1.6)} & 87.8 \textcolor{red}{(+8.6)} & 87.3 \textcolor{red}{(+8.1)} & 90.8 \textcolor{red}{(+11.6)} & 91.5 \textcolor{red}{(+12.3)} & 91.2 \textcolor{red}{(+12.0)} & 91.5 \textcolor{red}{(+12.3)} & 91.9 \textcolor{red}{(+12.7)} \\
    MMMU & 48.4 & 51.4 \textcolor{red}{(+3.0)} & 54.2 \textcolor{red}{(+5.8)} & 58.5 \textcolor{red}{(+10.1)} & 54.8 \textcolor{red}{(+6.4)} & 59.2 \textcolor{red}{(+10.8)} & 60.8 \textcolor{red}{(+12.4)} & 61.3 \textcolor{red}{(+12.9)} & 61.1 \textcolor{red}{(+12.7)} & 61.6 \textcolor{red}{(+13.2)} \\
    MathVista & 53.0 & 51.2 \textcolor{blue}{(-1.8)} & 53.6 \textcolor{red}{(+0.6)} & 53.8 \textcolor{red}{(+0.8)} & 49.3 \textcolor{blue}{(-3.7)} & 52.3 \textcolor{blue}{(-0.7)} & 53.8 \textcolor{red}{(+0.8)} & 54.2 \textcolor{red}{(+1.2)} & 54.5 \textcolor{red}{(+1.5)} & 54.9 \textcolor{red}{(+1.9)} \\
    \bottomrule
\end{tabular}
}
\end{table*}

\begin{table*}[!t]
\caption{
Different model configurations of Textual Sub-Agents for implementing LLM-assisted visual reasoning in \name{}.
The abbreviations \texttt{4o-mini}, \texttt{Qwen72B}, and \texttt{Qwen3} refer to \texttt{GPT-4o-mini}, \texttt{Qwen2.5-72B-Instruct}, and \texttt{Qwen3-32B}, respectively. 
"Assisted" stands for LLM-assisted reasoning. 
}
\label{table:text agent set}
\resizebox{\textwidth}{!}{
\centering
\begin{tabular}{l l l l l l}
    \toprule
    \multirow{1}{*}{\begin{tabular}{c} Model \end{tabular}}
         &Method & Vision Expert & Summarizer & Other Textual Sub-Agents\\
    \midrule
    \multirow{2}{*}{\begin{tabular}{c} \texttt{Llama3-LLaVA-NeXT-8B} \end{tabular}}
         &\name{} + \texttt{4o-mini} Assisted & \texttt{Llama3-LLaVA-NeXT-8B} & \texttt{GPT-4o-mini} & \texttt{GPT-4o-mini}\\ 
         &\name{} + \texttt{Qwen72B} Assisted & \texttt{Llama3-LLaVA-NeXT-8B} & \texttt{Qwen2.5-72B-Instruct} & \texttt{Qwen2.5-72B-Instruct}\\ 
    \midrule
        \multirow{2}{*}{\texttt{Qwen2.5-VL-7B-Instruct}} 
         &\name{} + \texttt{Qwen72B} Assisted & \texttt{Qwen2.5-72B-Instruct} & \texttt{Qwen2.5-72B-Instruct} & \texttt{Qwen2.5-72B-Instruct}\\ 
         &\name{} + \texttt{Qwen3} Assisted & \texttt{Qwen2.5-72B-Instruct} & \texttt{Qwen3-32B} & \texttt{Qwen2.5-72B-Instruct}\\ 
    \bottomrule
    \end{tabular}
    }
\end{table*}

\clearpage
\clearpage
\newpage
\onecolumn
\subsection{Demonstrative Examples}

\begin{figure*}[h!]
\centering
\includegraphics[width=0.88\linewidth]{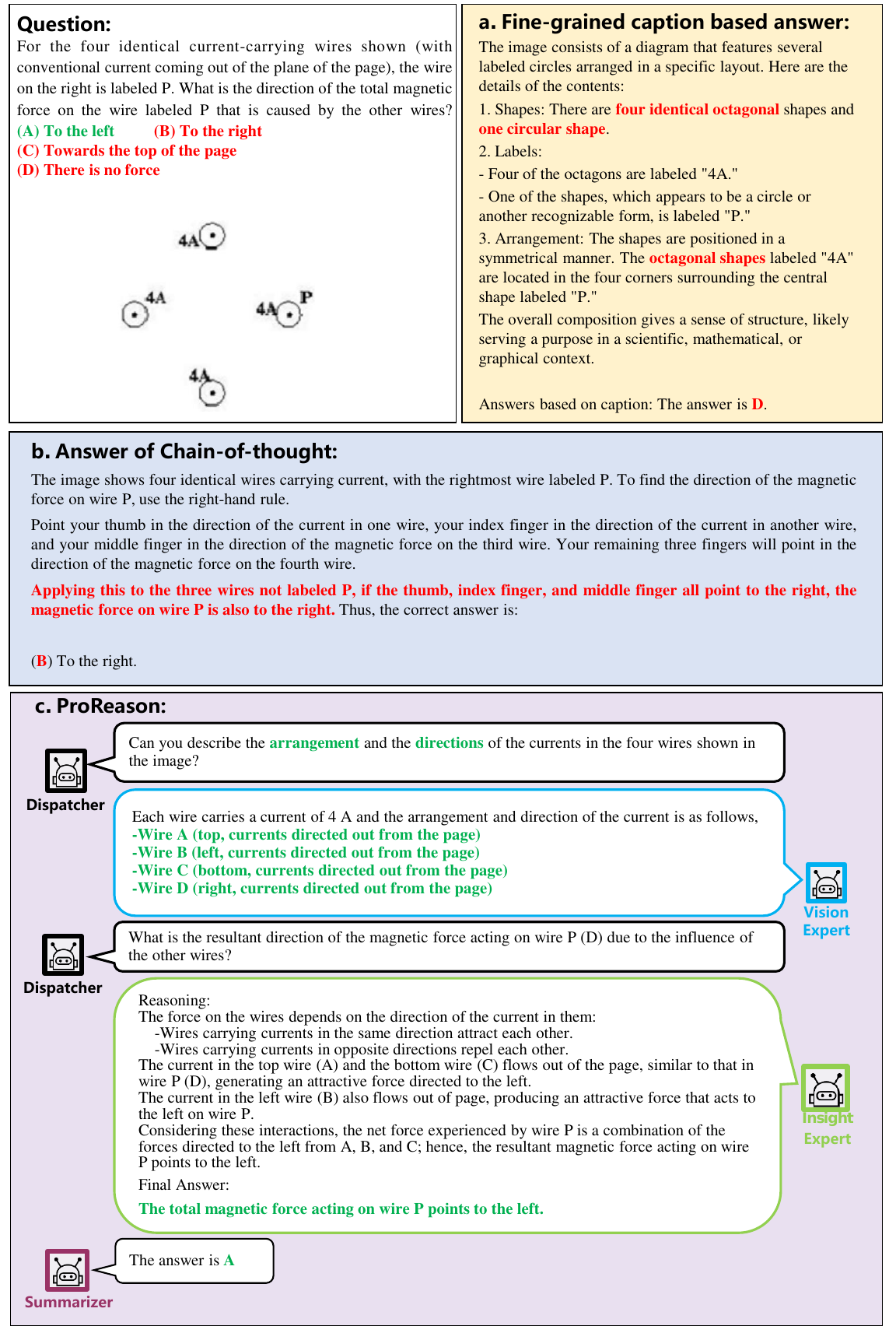}
\caption{An example with three reasoning frameworks: fine-grained caption, chain-of-thought, and \name{}.
\name{} enables LVLMs to proactively acquire necessary information in a question-oriented manner, and predicts answers based on the collected information. Apparently, \name{} is superior to previous methods, which often describe question-irrelevant visual details, or overlook informative elements.
Green indicates correct information or conclusions, while red signifies incorrect ones.}
\label{fig:case}
\end{figure*}

\begin{figure*}[t]
\centering
\includegraphics[width=\linewidth]{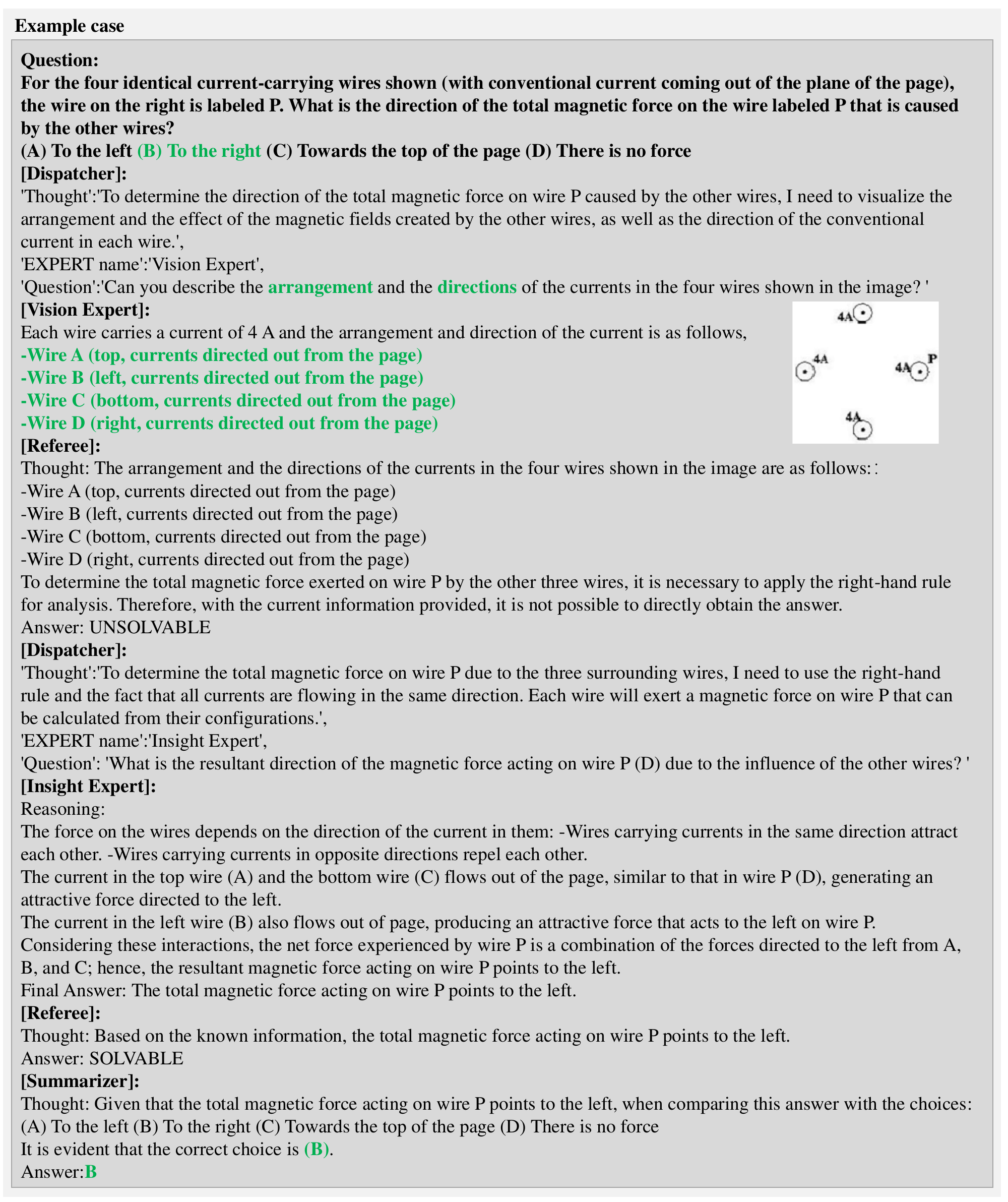}
\caption{A complete reasoning process of \name{} for the case shown in Figure~\ref{fig:case}.}
\label{fig:full case}
\end{figure*}

\begin{figure*}[t]
\centering
\includegraphics[page=1, width=\linewidth]{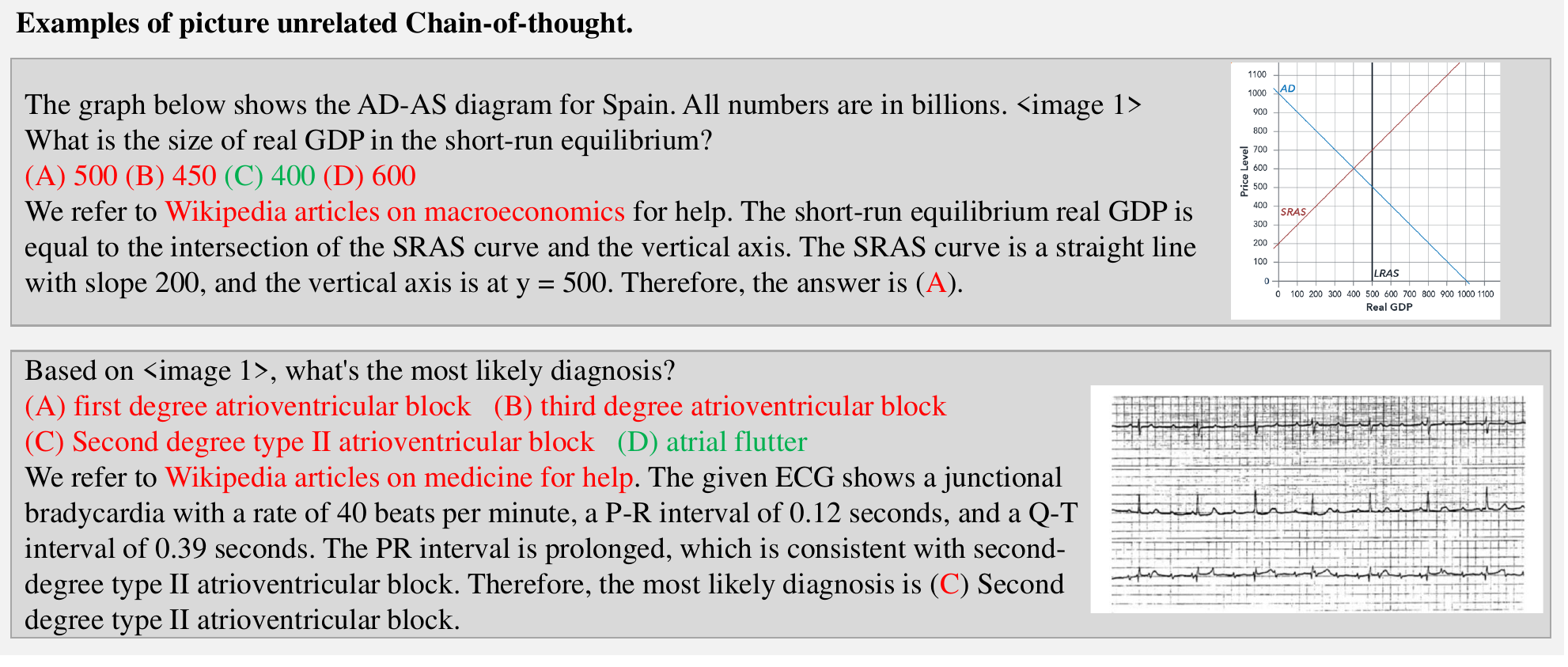}
\caption{
Additional examples of images-unrelated Chain-of-Thought reasoning.
}
\label{fig:example_cot}


\centering
\includegraphics[page=2, width=\linewidth]{example.pdf}
\caption{
Additional examples of question-unrelated passive image captions.
}
\label{fig:example_cap}
\end{figure*}

\begin{figure*}[b]
\centering
\includegraphics[page=1, width=\linewidth]{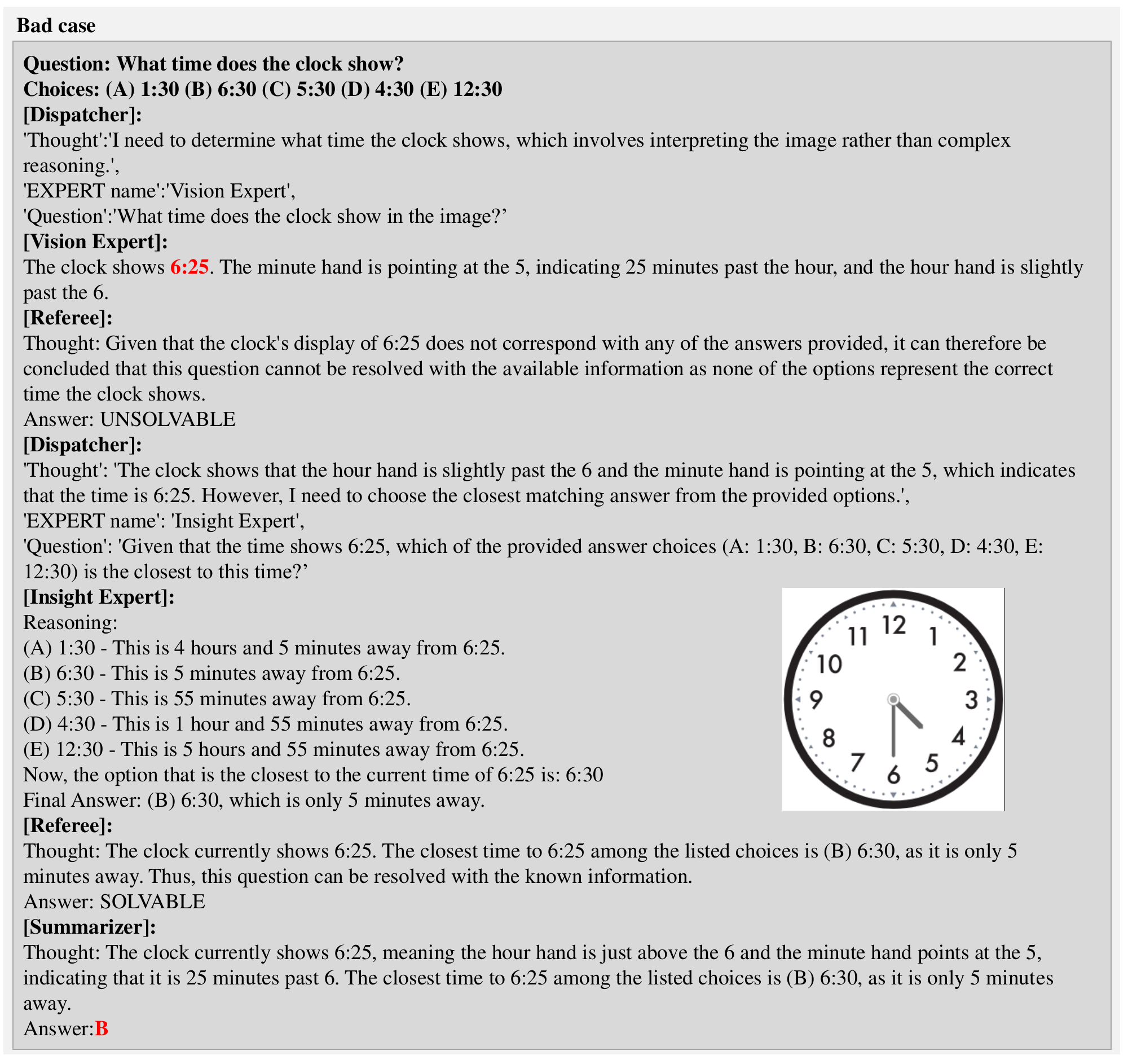}
\caption{
A typical mistake made by \name{}.
The vision expert incorrectly identifies 4:30 as 6:25, leading the other agents to base their judgments on this erroneous information, and ultimately resulting in the wrong conclusion.
}
\label{fig:error case1}
\end{figure*}

\begin{figure*}[h]
\centering
\includegraphics[page=2, width=\linewidth]{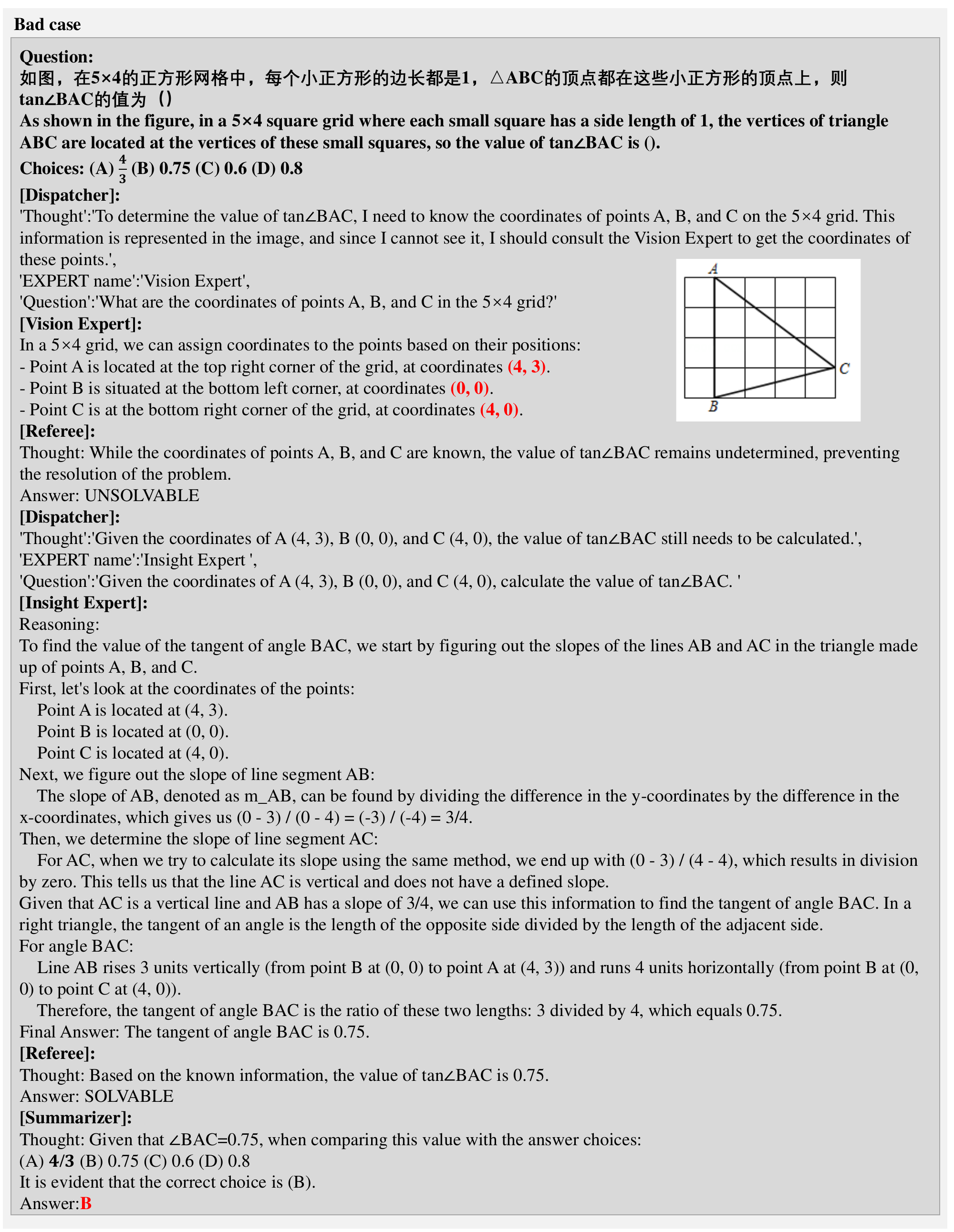}
\caption{
A typical mistake made by \name{}.
The vision expert incorrectly provided the coordinates of the three vertices of the triangle, leading the other agents to base their judgments on this erroneous information, and ultimately resulting in the wrong conclusion.
}
\label{fig:error case2}
\end{figure*}

\clearpage
\clearpage
\newpage
\onecolumn
\subsection{Prompts}

\begin{figure}[h]
\centering
\includegraphics[width=\linewidth]{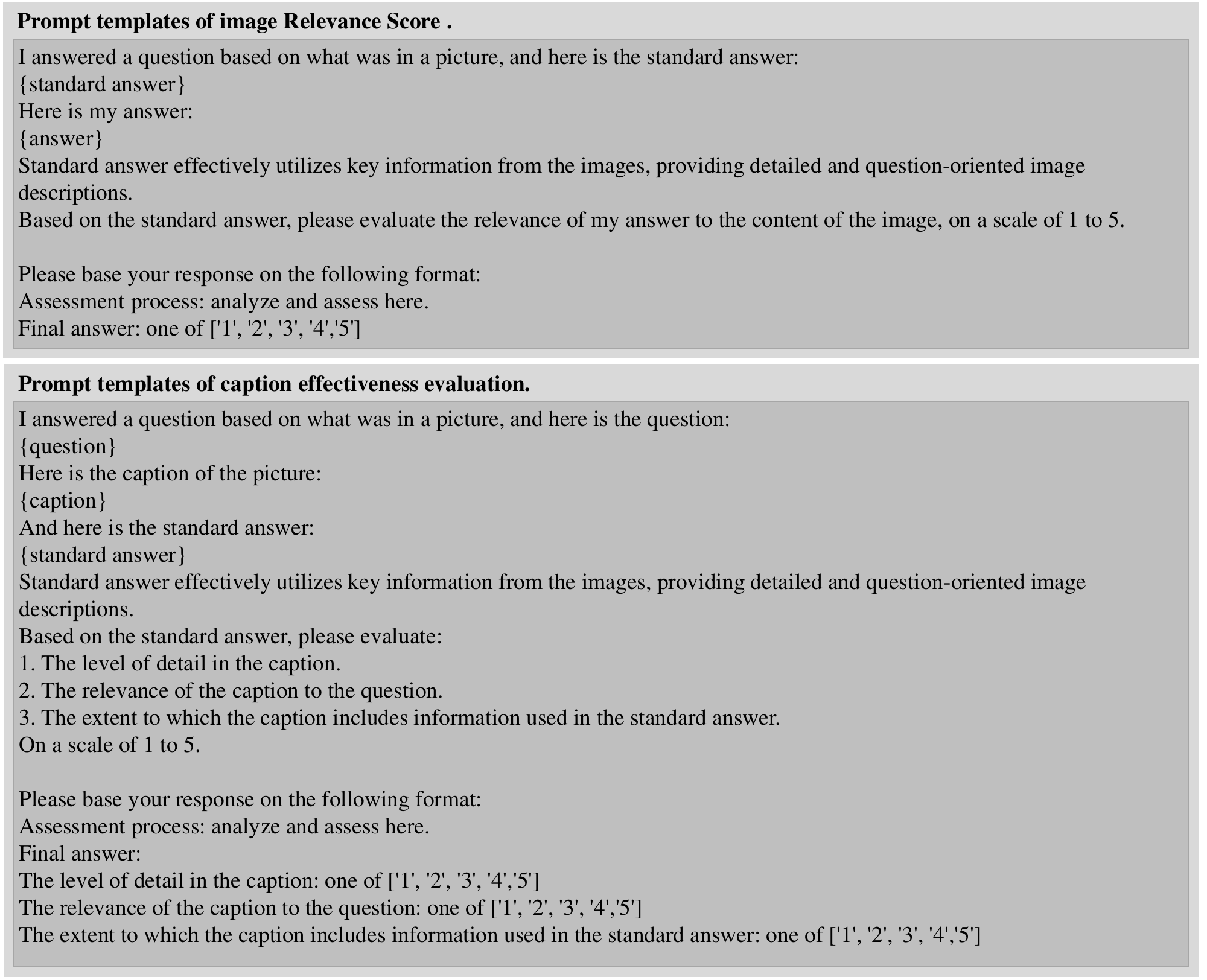}
\caption{Prompt templates of Relevance Score and caption effectiveness evaluation.}
\label{fig:score}
\end{figure}

\newpage
\twocolumn
\begin{figure*}[h]
\centering
\includegraphics[width=\linewidth]{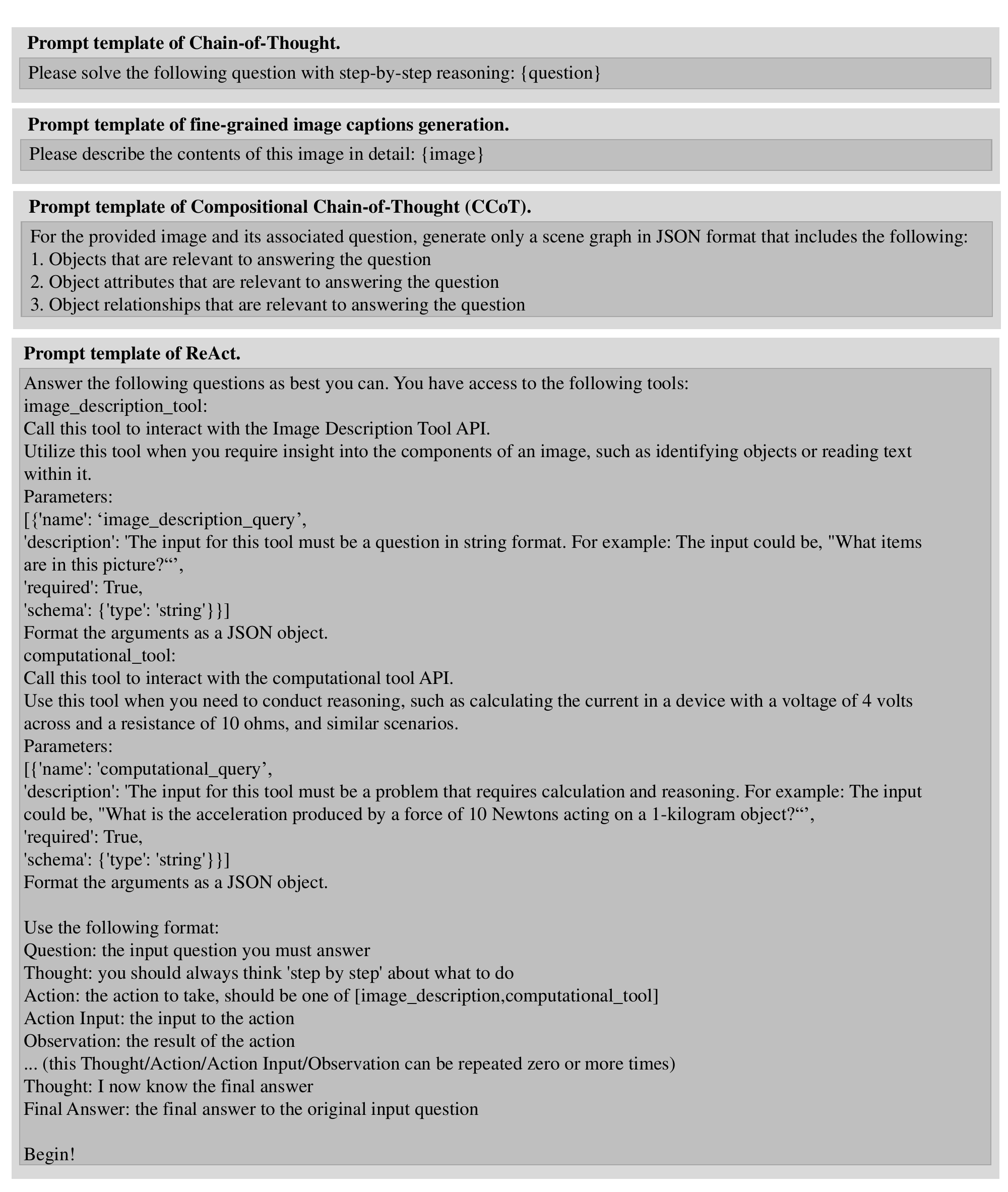}
\caption{Prompt templates of Chain-of-Thought, fine-grained image captions generation, Compositional Chain-of-Thought (CCoT), and ReAct.}
\label{fig:prompt_baseline}
\end{figure*}

\begin{figure*}[b]
\centering
\includegraphics[width=\linewidth]{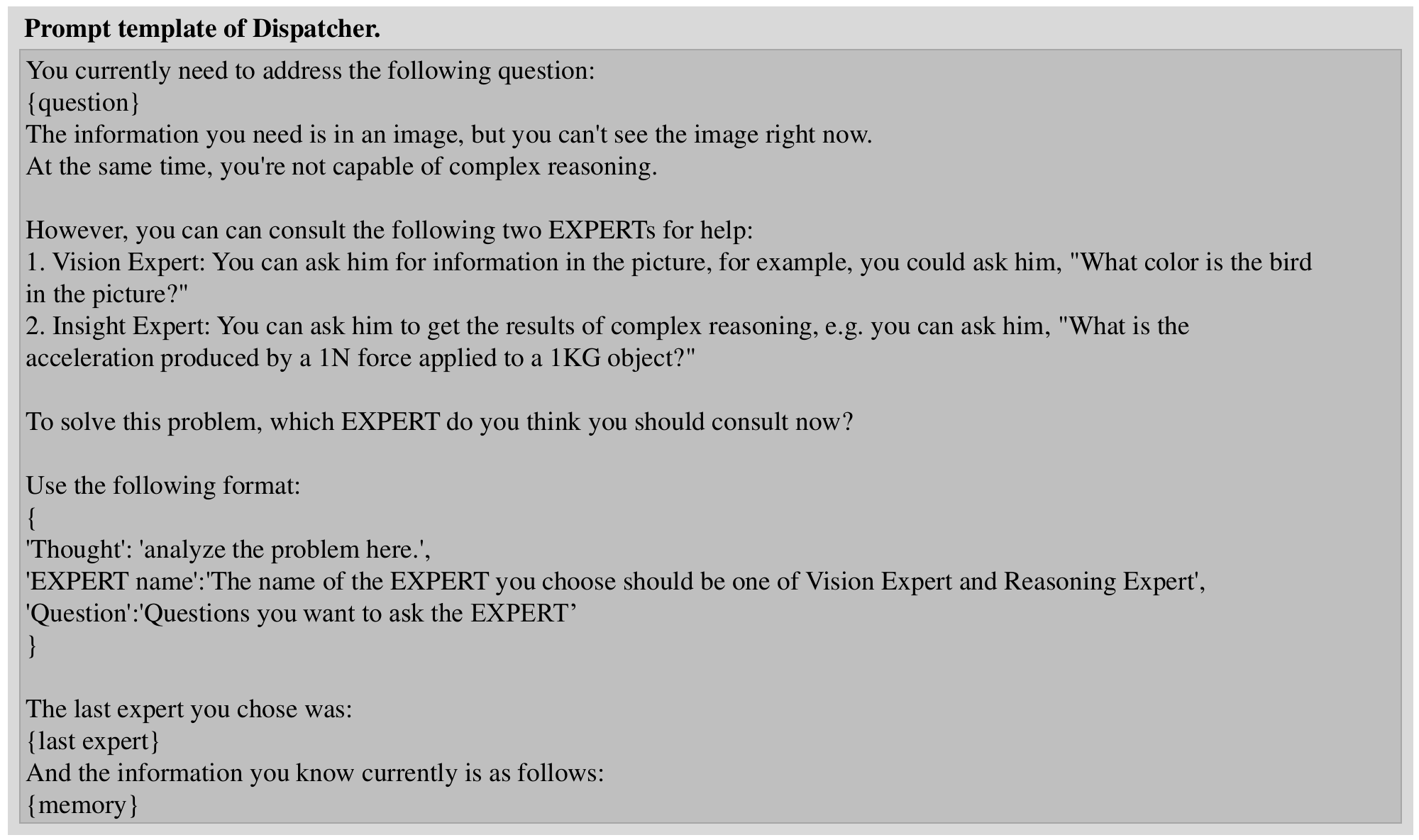}
\caption{Prompt templates of Dispatcher.}
\label{fig:prompt_dis}
\end{figure*}

\end{document}